\documentclass[10pt,twocolumn,letterpaper]{article}

\usepackage{cvpr}              

\usepackage{graphicx}
\usepackage{amsmath}
\usepackage{amssymb}
\usepackage{amsthm}
\usepackage{booktabs}
\usepackage{mathrsfs} 
\usepackage{bm}
\usepackage{bbm}
\usepackage{algorithm}
\usepackage{algorithmicx}
\usepackage{algpseudocode}
\usepackage{multirow}
\usepackage{color}
\usepackage{wrapfig}
\usepackage{subcaption}
\usepackage[pagebackref,breaklinks,colorlinks]{hyperref}
\usepackage[hoptionsi]{overpic}
\usepackage[table]{xcolor}
\definecolor{lightgrey}{rgb}{0.92,0.92,0.92}
\usepackage{xcolor}
\usepackage[normalem]{ulem}
\usepackage{balance}

\usepackage[capitalize]{cleveref}
\crefname{section}{Sec.}{Secs.}
\Crefname{section}{Section}{Sections}
\Crefname{table}{Table}{Tables}
\crefname{table}{Tab.}{Tabs.}

\def\cvprPaperID{1739}
\def\confName{CVPR}
\def\confYear{2023}

\newtheorem{theorem}{Theorem}

\newcommand{\name}{\textcolor{black}{LiVT\ }}
\newcommand{\good}[1]{\textcolor[RGB]{55, 146, 55}{\small {\textbf{#1}}}}
\newcommand{\bad}[1]{\textcolor[RGB]{183, 62, 62}{\small {\textbf{#1}}}}

\newcommand\hl{\bgroup\markoverwith{\textcolor[RGB]{210,210,210}{\rule[-.5ex]{2pt}{2.5ex}}}\ULon}

\begin{document}

\title{Learning Imbalanced Data with Vision Transformers}

\author{Zhengzhuo Xu ~~~~~ Ruikang Liu ~~~~~ Shuo Yang ~~~~~ Zenghao Chai ~~~~~ Chun Yuan{$^\dagger$}\\
Shenzhen International Graduate School, Tsinghua University, China\\
{\tt\small \{xzzthu, liuruikang.cs, ysss9264, zenghaochai\}@gmail.com ~~ yuanc@sz.tsinghua.edu.cn}
}

\maketitle
\begin{abstract}
 The real-world data tends to be heavily imbalanced and severely skew the data-driven deep neural networks, which makes Long-Tailed Recognition (LTR) a massive challenging task. 
Existing LTR methods seldom train Vision Transformers (ViTs) with Long-Tailed (LT) data, while the off-the-shelf pretrain weight of ViTs always leads to unfair comparisons.
In this paper, we systematically investigate the ViTs' performance in LTR and propose \name to train ViTs \textbf{from scratch} only with LT data.
With the observation that ViTs suffer more severe LTR problems, we conduct Masked Generative Pretraining (MGP) to learn generalized features. 
With ample and solid evidence, we show that MGP is more robust than supervised manners.
Although Binary Cross Entropy (BCE) loss performs well with ViTs, it struggles on the LTR tasks.
We further propose the balanced BCE to ameliorate it with strong theoretical groundings. 
Specially, we derive the unbiased extension of Sigmoid and compensate extra logit margins for deploying it.
Our Bal-BCE contributes to the quick convergence of ViTs in just a few epochs.
Extensive experiments demonstrate that with MGP and Bal-BCE, \name successfully trains ViTs well without any additional data and outperforms comparable state-of-the-art methods significantly, e.g., our ViT-B achieves 81.0\% Top-1 accuracy in iNaturalist 2018 without bells and whistles. 
Code is available at \url{https://github.com/XuZhengzhuo/LiVT}.
\end{abstract}
\section{Introduction}

With the vast success in the computer vision field, \textbf{Vi}sion \textbf{T}ransformers (ViTs) \cite{ViT, Swin} get increasingly popular and have been widely used in visual recognition \cite{ViT}, detection \cite{DETR}, and video analysis \cite{MASL}. These models are heavily dependent on large-scale and balanced data to avoid overfitting \cite{PlaceLT, Imagenet, COCO}. However, real-world data usually confronts severe class-imbalance problems, \textit{i.e.}, most labels (tail) are associated with limited instances while a few categories (head) occupy dominant samples. The models simply classify images into head classes for lower error because the head always overwhelms tail ones in LTR. The data paucity also results in the model overfitting on the tail with unaccepted generalization. The aforementioned problems make \textbf{L}ong \textbf{T}ail \textbf{R}ecognization (LTR) a challenging task.

\begin{figure}[t]
\flushleft 
\begin{overpic}[scale=0.40,grid=False,tics=5]{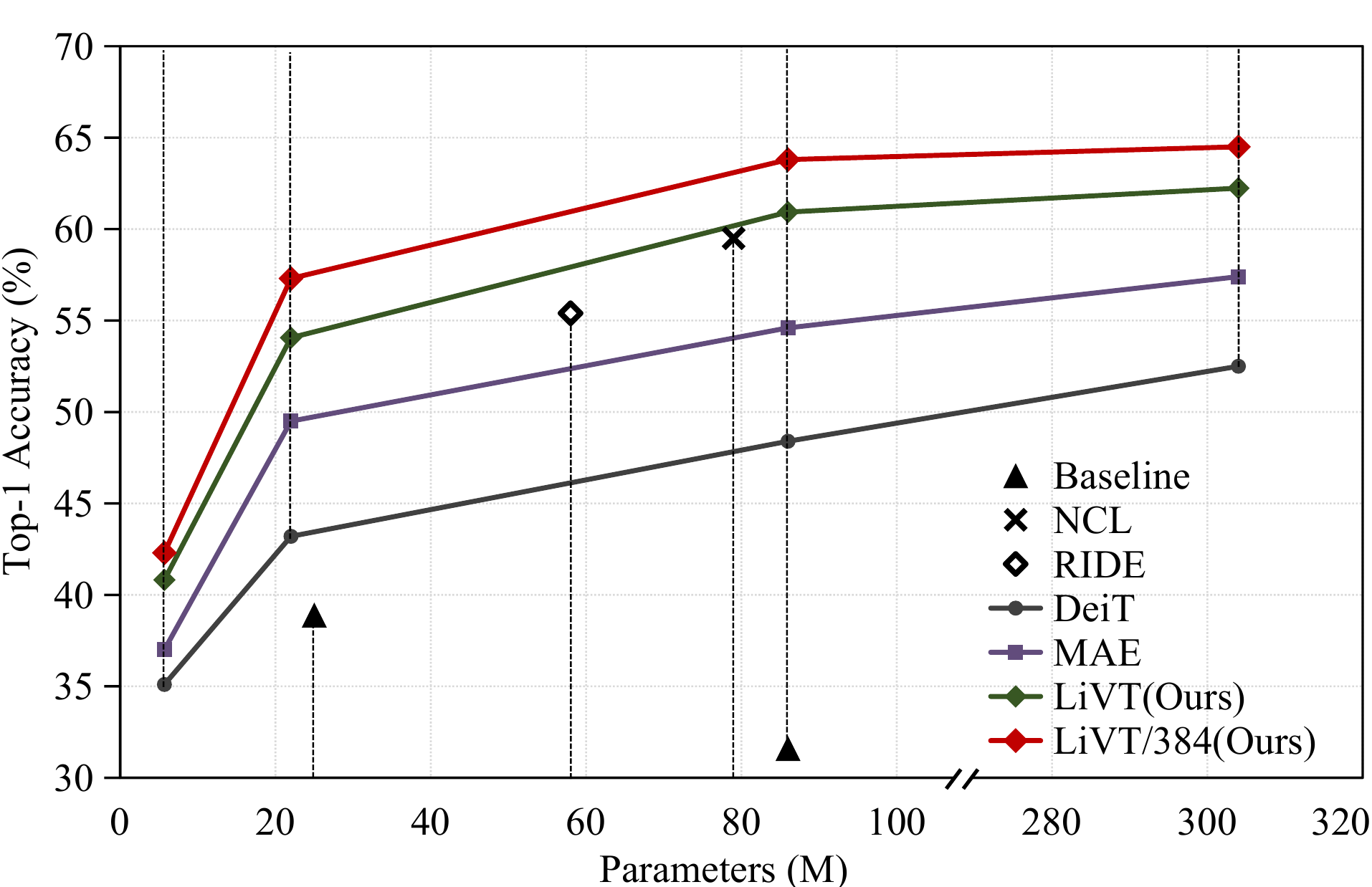}
\put(5,62){\scriptsize{ViT-Tiny}}
\put(17,62){\scriptsize{ViT-Small}}
\put(52,62){\scriptsize{ViT-Base}}
\put(85,62){\scriptsize{ViT-Large}}
\put(20,12){\scriptsize{R50}}
\put(36,13){\scriptsize{RIDE-4E}}
\put(39,10){\scriptsize{R50}}
\put(50,23){\scriptsize{NCL}}
\put(48,20){\scriptsize{R50$\times$3}}
\put(86.5,29){\scriptsize{\cite{CB}}}
\put(82,25.7){\scriptsize{\cite{NCL}}}
\put(84,22.5){\scriptsize{\cite{RIDE}}}
\put(84,19){\scriptsize{\cite{deit3}}}
\put(84,16){\scriptsize{\cite{MAE}}}
\end{overpic}
\caption{Top-1 Acc \textit{v.s.} Model Size on ImageNet-LT dataset. We choose the Tiny / Small / Base / Large ViT and multi-expert approaches. R50 represents the ResNet50 model. ViT-Base gets lower Acc than ResNet50 when trained in a supervised manner.}
\label{fig:size-acc}
\end{figure}
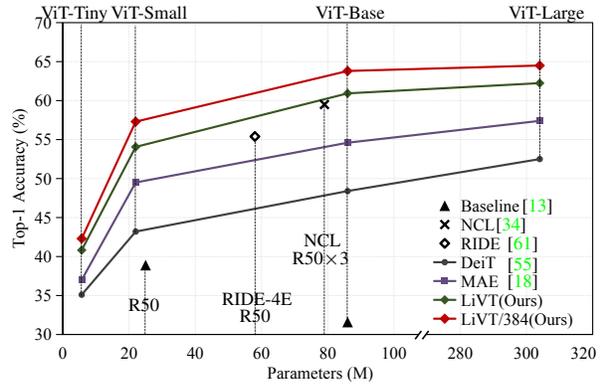

Numerous papers \cite{CB, OLTR, LDAM, LADE, PriorLT, NCL, GCL} handle the LTR problem with traditional supervised cross-entropy learning based on ResNet \cite{ResNet} or its derivatives \cite{ResNeXt}. Some methods use ViTs with pretrained weights on ImageNet \cite{Imagenet} (or larger datasets), which leads to unfair comparisons with additional data, \textit{e.g.} on ImageNet-LT (a subset of ImageNet-1K) benchmark. Moreover, there are still limited explorations on the utilization of \textbf{L}ong-\textbf{T}ailed (LT) data to train ViTs effectively. Therefore, in this paper, \textit{we try to train ViTs from scratch with LT data}. We observe that it is particularly difficult to train ViT with LT labels' supervision. As Tab.~\ref{tab:pretrain-recips} shows, ViTs degrade heavily when training data become skewed. ViT-B is much worse than ResNet50 with the same CE training manner (\textit{c.f.} Fig.~\ref{fig:size-acc}). One reasonable explanation is that ViTs require longer training to learn the inductive bias, while CNNs offer the built-in translation invariance implicitly. Yet another one lies in the label statistical bias in the LTR datasets, which confuses models to make predictions with an inherent bias to the head \cite{LA, PaCo}. The well-trained ViTs have to overcome the above plights simultaneously to avoid falling into dilemmas.

Inspired by decoupling \cite{NCM}, many methods \cite{MiSLAS, Remix, HybirdSC, PaCo, BCL} attempt to enhance feature extraction in supervised manners like mixup \cite{mixup} / remix \cite{Remix}, or \textbf{S}elf-\textbf{S}upervised \textbf{L}earning (SSL) like \textbf{C}ontrastive \textbf{L}earning (CL) \cite{SimCLR, MoCo}. Liu \textit{et al.} \cite{rwSAM} claim that SSL representations are more robust to class imbalance than supervised ones, which inspires us to train ViTs with SSL. However, CL is quite challenging for extensive memory requisition and converge difficulties \cite{MoCo3}, where more explorations are required to work well with ViTs in LTR. In contrast, we propose to \textbf{L}earn \textbf{i}mbalanced data with \textbf{V}i\textbf{T}s ({LiVT}) by \textbf{M}asked \textbf{G}enerative \textbf{P}retraining (MGP) and \textbf{B}alanced \textbf{F}ine \textbf{T}uning (BFT).

Firstly, LiVT adopts MGP to enhance ViTs' feature extraction, which has been proven effective on BeiT \cite{BeiT} and MAE \cite{MAE}. It reconstructs the masked region of images with an extra lightweight decoder. We observe that MGP is stable with ViTs and robust enough to LT data with empirical evidence. Despite the label distribution, the comparable number of training images will bring similar feature extraction ability, which greatly alleviates the toxic effect of LT labels \cite{DAP}. Meanwhile, the training is accelerated by masked tokens with acceptable memory requisition.

Secondly, LiVT trains the downstream head with rebalancing strategies to utilize annotation information, which is consistent with \cite{NCM, MiSLAS, GCL}. Generally, \textbf{B}inary \textbf{C}ross-\textbf{E}ntropy (BCE) loss performs better than Cross-Entropy loss when collaborating with ViTs \cite{deit3}. However, it fails to catch up with widely adopted \textbf{Bal}anced \textbf{C}ross-\textbf{E}ntropy (Bal-CE) loss and shows severe training instability in LTR. We propose the \textbf{Bal}anced \textbf{BCE} (Bal-BCE) loss to revise the mismatch margins given by Bal-CE. Detailed and solid theoretical derivations are provided from Bayesian theory. Our Bal-BCE ameliorates BCE by a large margin and achieves state-of-the-art (SOTA) performance with ViTs.

Extensive experiments show that LiVT learns LT data more efficiently and outperforms vanilla ViT \cite{ViT}, DeiT III \cite{deit3}, and MAE \cite{MAE} remarkably. As detailed comparisons in Fig.~\ref{fig:size-acc}, LiVT achieves SOTA on ImageNet-LT with affordable parameters, despite that ImageNet-LT is a relatively small dataset for ViTs. The ViT-Small \cite{deit3} also achieves outstanding performance compared to ResNet50. Our key contributions are summarized as follows.
\vspace{-5pt}
\begin{itemize}
    \setlength\itemsep{-0.2em}
    \item To our best knowledge, we are the first to investigate training ViTs from scratch with LT data systematically.
    \item We pinpoint that the masked generative pretraining is robust to LT data, which avoids the toxic influence of imbalanced labels on feature learning.
    \item With a solid theoretical grounding, we propose the balanced version of BCE loss (Bal-BCE), which improves the vanilla BCE by a large margin in LTR. 
    \item We propose LiVT recipe to train ViTs from scratch, and the performance of LiVT achieves state-of-the-art across various benchmarks for long-tailed recognition.
\end{itemize}

\begin{table}[t]
\centering
\caption{Top-1 accuracy (\%) of different recipes to train ViT-B-16 from scratch on ImageNet-LT/BAL. All perform much worse on LT than BAL. See descriptions of LT \& BAL in section \ref{sec:dataset}.}
\resizebox{1.0\linewidth}{8.4mm}{
\begin{tabular}{@{}c|cc|cc|cc@{}}
\toprule
Dataset      & ViT  & $\Delta$ & DeiT III & $\Delta$ & MAE  & $\Delta$ \\ \midrule
ImageNet-BAL & 38.7 & -    & 67.2     & -     & 69.2 & -     \\
ImageNet-LT  & 31.6 & -7.0 & 48.4     & -18.8 & 54.5 & -14.7 \\ \bottomrule
\end{tabular}}
\label{tab:pretrain-recips}
\end{table}
\section{Related Work}

\subsection{Long-tailed Visual Recognition}
We roughly divide LTR progress into three groups.

\noindent\textbf{Rebalancing strategies} adjust each class contribution with delicate designs. Re-sampling methods adopt class-wise sampling rate to learn balanced networks \cite{BBN, CB, CreST, ID-samlping, GCL}. More sophisticated approaches replenish few-shot samples with the help of many-shot ones \cite{M2m, FSA, PriorLT, Remix, Bagoftricks, CMO}. The re-weighting proposals modify the loss function by adjusting class weights \cite{Focal, CB, EQL, CausalNorm, IB, MiSLAS, LTR-WD} to assign different weights to samples or enlarging logit margins \cite{LDAM, LA, BS, PriorLT, LADE, TADE, DisAlign, GCL} to learn more challenging and sparse classes. However, the rebalancing strategies are always at the cost of many-shot accuracy inevitably.

\noindent\textbf{Multi-Expert networks} alleviate the LTR problem with \textit{single expert learning} and \textit{knowledge aggregation }\cite{BBN, LFME, RIDE, CBD, TADE, ACE, DiVE, SSD, TLC, NCL}. LFME \cite{LFME} trains experts with the subsets with a lower imbalance ratio and aggregate via knowledge distillation. TADE \cite{TADE} learns three classifiers with the different test labels prior based on Logit Adjustment \cite{LA} and optimizes classifiers' output weights by contrastive learning \cite{SimCLR}. NCL \cite{NCL} collaboratively learns multiple experts together to reduce tail uncertainty. However, it is still heuristic to design expert individual training and knowledge aggregation manners. The overly complex models also make training difficult and limit the inference speed.

\noindent\textbf{Multi-stage training} is another effective training strategy for LTR. Cao \textit{et al.}\cite{LDAM} propose to learn features at first and defer re-weighting in the second stage. Kang \textit{et al.} \cite{NCM} further decouples the representation and classifier learning separately, where the classifier is trained with re-balancing strategies just in the second stage. Some works \cite{MiSLAS, PriorLT, Remix} adopt more approaches, \textit{e.g.}, mixup \cite{mixup} or remix \cite{Remix}, to improve features in the first stage. More recently, \textbf{C}ontristive \textbf{L}earning (CL)~\cite{SimCLR, MoCo} is gaining increasing concern. Kang \textit{et al.} \cite{EBFS} exploit to learn balanced feature representations by CL to bypass the influence of imbalanced labels. However, it is more effective to adopt \textbf{S}upervised \textbf{C}ontristive \textbf{L}earning (SCL) to utilize the labels \cite{SSP, HybirdSC}. With SCL, SOTAs \cite{PaCo, KCL, TSC, BCL} all adopt the Bal-CE loss \cite{BS, LA, LADE, PriorLT} to train the classifier for better performance. Masked Generative learning \cite{BERT, MaskGIT, MAE} is another effective feature learning method. However, there is still limited research on it in the community of LTR.

\subsection{Vision Transformers}
Current observations and conclusions are mostly based on ResNets \cite{ResNet, ResNeXt}. Most recently, ViT \cite{ViT} has shown extraordinary performance after pre-training on large-scale and balanced datasets. Swin transformer \cite{Swin} proposes a hierarchical transformer with shift windows to bring greater efficiency. DeiT \cite{deit3} introduces a simple but effective recipe to train ViT with limited data. BeiT \cite{BeiT} trains ViT with the idea of Mask Language Models. MAE \cite{MAE} further reduces the computation complexity with a lightweight decoder and higher mask ratio. Although RAC \cite{RAC} adopts ViTs with pretrained checkpoints, there is limited research to train ViTs from scratch on long-tailed datasets.
\section{Preliminaries}

\subsection{Task Definition}
With a $N$-sample and $C$-class dataset $\mathcal{D}=\{\mathcal{X}, \mathcal{Y}\}$, we note each instance $\mathbf{x}_i \in \mathcal{X} := \{\mathbf{x}_1,\cdots,\mathbf{x}_N\}$ and corresponding $\mathbf{y}_i \in \mathcal{Y}:=\{\mathbf{y}_1,\cdots,\mathbf{y}_N\}$, where each $\mathbf{y}_i \in \mathcal{C}:=\{1,\cdots,C\}$. In long-tailed visual recognition, each category $\mathcal{C}_i$ has a  different instance number $n_i=|\mathcal{C}_i|$ and we set $\gamma=n_{max}/n_{min}$ to measure how skewed the long-tailed dataset is. We train the model $\mathcal{M}:=\{\mathscr{F}_{\theta_f}, \mathscr{W}_{\theta_w}\}$ with $\mathcal{D}$, which contains a \textit{feature encoder} $\mathscr{F}_{\theta_f}$ and a \textit{classifier} $\mathscr{W}_{\theta_w}$. Besides, we consider a lightweight decoder $\mathscr{D}_{\theta_d}$ for mask autoencoder architecture. For an input image $\mathbf{x}$, the encoder extracts the feature representation $\mathbf{v}:=\mathscr{F}(\mathbf{x}|{\theta_f})\in \mathbb{R}^d$, the classifier gives the logits $\mathbf{z}:=\mathscr{W}(\mathbf{v}|{\theta_w}) \in \mathbb{R}^C$ and the decoder reconstructs original image $\hat{\mathbf{x}}:=\mathscr{D}(\mathbf{v}|{\theta_d}) \in \mathbb{R}^{H\times W\times 3}$. The $d$ / $H$ / $W$ is feature dimension / resized height / resized width, respectively.

\subsection{Balanced Cross-entropy}
Here, we revisit the balanced softmax and corresponding \textbf{Bal}anced \textbf{C}ross-\textbf{E}ntropy (BalCE) loss \cite{LA, BS, LADE, PriorLT, NCL, BCL}, which has been widely adopted in LTR. Consider the standard \textit{softmax} operation and cross-entropy loss:
\begin{equation}
\label{eq:ce}
    \begin{aligned}
        &\mathcal{L}_{\text{CE}}(\mathcal{M}(\mathbf{x}|\theta_f,\theta_w), \mathbf{y}_i) = - \log(p(\mathbf{y}_i|\mathbf{x};\theta_f,\theta_w)) \\
        &= - \log[e^{z_{\mathbf{y}_i}} / \sum_{\mathbf{y}_j \in \mathcal{Y}} e^{z_{\mathbf{y}_j}}] = \log[1+\sum_{\mathbf{y}_j\neq \mathbf{y}_i} e^{\mathbf{z}_{\mathbf{y}_j} - \mathbf{z}_{\mathbf{y}_i}}].
    \end{aligned}
\end{equation}

If we take the class instance number $n_{\mathbf{y}_i}$ into account for softmax \cite{BS}, we have the balanced cross-entropy loss:
\begin{equation}
\label{eq:bal-ce}
    \begin{aligned}
        &\mathcal{L}_{\text{Bal-CE}}(\mathcal{M}(\mathbf{x}|\theta_f,\theta_w), \mathbf{y}_i) = - \log(p(\mathbf{y}_i|\mathbf{x};\theta_f,\theta_w)) \\
        &= - \log[\frac{n_{\mathbf{y}_i} e^{z_{\mathbf{y}_i}}}{\sum_{\mathbf{y}_j \in \mathcal{Y}} n_{\mathbf{y}_j} e^{z_{\mathbf{y}_j}}}] \\
        &= \log[1+\sum_{\mathbf{y}_j\neq \mathbf{y}_i} e^{\log n_{\mathbf{y}_j} - \log n_{\mathbf{y}_i}} \cdot e^{\mathbf{z}_{\mathbf{y}_j} - \mathbf{z}_{\mathbf{y}_i}}].
    \end{aligned}
\end{equation}

\begin{theorem}
\label{thm:01_balance_softmax}
\normalfont{\hl{Logit Bias of Balanced CE.}} Let $\pi_{\mathbf{y}_i}=n_{\mathbf{y}_i} / N$ be the training label $\mathbf{y}_i$ distribution. If we implement the balanced cross-entropy loss via logit adjustment, the bias item of logit 
$\mathbf{z}_{\mathbf{y}_i}$ will be $\mathcal{B}^{\text{ce}}_{\mathbf{y}_i}=\log \pi_{\mathbf{y}_i}$, i.e.,
\end{theorem}
\begin{equation}
  \begin{aligned}
        \mathcal{L}_{\text{Bal-CE}} &= \log[1+\sum_{\mathbf{y}_j\neq \mathbf{y}_i} e^{\log n_{\mathbf{y}_j} - \log n_{\mathbf{y}_i}} \cdot e^{\mathbf{z}_{\mathbf{y}_j} - \mathbf{z}_{\mathbf{y}_i}}] \\
        &= \log[1+\sum_{\mathbf{y}_j\neq \mathbf{y}_i} e^{(\mathbf{z}_{\mathbf{y}_j} + \log n_{\mathbf{y}_j}) - (\mathbf{z}_{\mathbf{y}_i} + \log n_{\mathbf{y}_i})}] \\
        &= \log[1+\sum_{\mathbf{y}_j\neq \mathbf{y}_i} e^{(\mathbf{z}_{\mathbf{y}_j} + \log \pi_{\mathbf{y}_j}) - (\mathbf{z}_{\mathbf{y}_i} + \log \pi_{\mathbf{y}_i})}].
\end{aligned}  
\end{equation}

\textbf{\textit{Proof.}} See subsection 5.1 from \cite{LA} or detail derivation in the Appendix from the Bayesian Theorem perspective.

Bal-CE loss strengthens the tail instance's contributions while suppressing bias to the head, which alleviates the LTR problem effectively. However, the $\mathcal{B}^{\text{ce}}_{\mathbf{y}_i}$ in Thm.\ref{thm:01_balance_softmax} fails to work well when collaborating with BCE, where More analysis is required to build a \textit{balanced} version BCE loss.

\section{Methodology}
In this section, we introduce our \name in two stages. In section~\ref{sec:pretrain}, we revisit the generative masked auto-encoder as our first stage. Then, we propose the novel balanced sigmoid and corresponding binary cross entropy to collaborate with ViTs in section~\ref{sec:finetune}. Eventually, we summarize our whole pipeline in section~\ref{sec:pipeline}.

\subsection{Masked Generative Pretraining}
\label{sec:pretrain}
Inspired by BeiT \cite{BeiT} and MAE \cite{MAE}, we pretrain feature encoder $\mathscr{F}_{\theta_f}$ via MGP for its training efficiency and label irrelevance. MGP trains the encoder parameters $\theta_f$ with high ratio masked images and reconstructs the original image by a lightweight decoder $\mathscr{D}_{\theta_d}$.
\begin{equation}
\hat{\mathbf{x}}=\mathscr{D}_{\theta_d}\left(\mathscr{F}_{\theta_f}(\mathbf{M} \odot \mathbf{x})\right),
\end{equation}
where $\mathbf{M}\in \{0,1\}^{H\times W}$ is a random patch-wise binary mask. Then, we optimize $\theta_f, \theta_d$ end-to-end via minimizing the mean squared error between $\mathrm{x}$ and $\hat{\mathbf{x}}$.

We adopt MGP for two reasons: 1) \textit{It is difficult to train ViTs directly with label supervision} (see plain ViT-B performance in Fig.~\ref{fig:size-acc}) for its convergence difficulty and computation requirement. The DeiT III~\cite{deit3} is hard to catch up with SOTAs in LTR, even with more training epochs, stronger data augmentation, and larger model sizes. 2) \textit{The feature extraction ability of MGP is affected slightly by class instance number}, compared with previous mixup-based supervision~\cite{NCM, MiSLAS, GCL}, CL~\cite{HybirdSC} or SCL~\cite{PaCo, KCL, TSC, BCL}. Even pretraining on LTR datasets, the transfer performance of MGP is on par with that trained on balanced datasets with comparable total training instances. See transfer results in Tab.~\ref{tab:performance-transfer} and more visualization in Appendix.

\subsection{Balanced Fine Tuning}
\label{sec:finetune}
In the Balanced Fine-Tuning (BFT) phase, \textit{softmax} + CE loss has been the standard paradigm for utilizing annotated labels. However, recent research \cite{timm, deit3, TokenMix} pinpoint that Binary Cross-Entropy (BCE) loss works much well with ViTs and is more convenient when employed with mixup-manners \cite{mixup, Cutmix, TokenMix}, which can be written as:
\begin{equation}
\label{eq:bce}
\begin{aligned}
    \mathcal{L}_{\text{BCE}} = - \sum_{\mathbf{y}_i \in \mathcal{C}} & w_{\mathbf{y}_i}[\mathbbm{1}(\mathbf{y}_i) \cdot \log \sigma(\mathbf{z}_{\mathbf{y}_i}) \\
    &+(1-\mathbbm{1}(\mathbf{y}_i)) \cdot \log (1-\sigma(\mathbf{z}_{\mathbf{y}_i}))],
\end{aligned}
\end{equation}
where $\sigma(x)=1/(1+e^{-x})$ indicates the \textit{sigmoid} operation.

In LTR, Balanced CE (Eq.~\ref{eq:bal-ce}) improves original CE (Eq.~\ref{eq:ce}) remarkably. However, we observe that it is not directly applicable when it comes to BCE. The logit bias $\mathcal{B}_{\mathbf{y}_i}$ in Thm.~\ref{thm:01_balance_softmax} leads to an even worse situation. Here, we claim that the proper bias of BCE shall be revised as Thm.~\ref{thm:02_balance_sigmoid} when collaborating with BCE in LTR.
\begin{theorem}
\label{thm:02_balance_sigmoid}
\normalfont{\hl{Logit Bias of Balanced BCE.}} Let $\pi_{\mathbf{y}_i}=n_{\mathbf{y}_i} / N$ be the class $\mathbf{y}_i$ distribution. If we implement the balanced binary cross-entropy loss via logit adjustment, the bias item of logit $\mathbf{z}_{\mathbf{y}_i}$ will be $\mathcal{B}^{\text{bce}}_{\mathbf{y}_i}=\log \pi_{\mathbf{y}_i} \bm{- \log (1-\pi_{\mathbf{y}_i})}$,
\end{theorem}
\vspace{-15pt}
\begin{equation}
\label{eq:balbce}
\begin{aligned}
    &\mathcal{L}_{\text{Bal-BCE}} = -\sum_{\mathbf{y}_i \in \mathcal{C}} w_{i}[\mathbbm{1}(\mathbf{y}_i) \cdot \log \frac{1}{1+e^{-[\mathbf{z}_{\mathbf{y}_i} + \log \pi_{\mathbf{y}_i} - \log (1-\pi_{\mathbf{y}_i})]}} \\
    &+(1-\mathbbm{1}(\mathbf{y}_i)) \cdot \log (1-\frac{1}{1+e^{-[\mathbf{z}_{\mathbf{y}_i} + \log \pi_{\mathbf{y}_i} - \log (1-\pi_{\mathbf{y}_i})]}})]
\end{aligned}
\end{equation}
\textbf{\textit{Proof.}} We regard Binary CE as $C$ binary classification loss. Hence, for the class $\mathbf{y}_i$, $\pi_{\mathbf{y}_i}$ indicates positive samples proportion and $1-\pi_{\mathbf{y}_i}$ indicates negative ones. Here, we start by revising the \textit{sigmoid} activation function:
\begin{equation}
\label{eq:sigmoid}
    \sigma(\mathbf{z}_{\mathbf{y}_i}) = \frac{1}{1 + e^{-\mathbf{z}_{\mathbf{y}_i}}}
    = \frac{e^0}{e^0 + e^{-\mathbf{z}_{\mathbf{y}_i}}}
    = \frac{e^{\mathbf{z}_{\mathbf{y}_i}}}{e^{\mathbf{z}_{\mathbf{y}_i}} + e^0}
\end{equation}

If we view Eq.~\ref{eq:sigmoid} as the binary version of \textit{softmax}, $e^x$ ($e^0$) will be the normalized probability to indicate \textit{yes} (\textit{no}). Similar to Eq.~\ref{eq:bal-ce}, we use instance number to balance \textit{sigmoid}:
\begin{equation}
\begin{aligned}
    \hat{\sigma}(\mathbf{z}_{\mathbf{y}_i}) &= \frac{n_{\mathbf{y}_i} \cdot e^{\mathbf{z}_{\mathbf{y}_i}}}{n_{\mathbf{y}_i} \cdot e^{\mathbf{z}_{\mathbf{y}_i}} + (N-n_{\mathbf{y}_i}) \cdot e^0} \\
    &= \frac{\pi_{\mathbf{y}_i} \cdot e^{\mathbf{z}_{\mathbf{y}_i}}}{\pi_{\mathbf{y}_i} \cdot e^{\mathbf{z}_{\mathbf{y}_i}} + (1-\pi_{\mathbf{y}_i}) \cdot e^0} \\
    &= \frac{1}{1 + \frac{1-\pi_{\mathbf{y}_i}}{\pi_{\mathbf{y}_i}} \cdot e^{-\mathbf{z}_{\mathbf{y}_i}}} \\
\end{aligned}
\end{equation}

Considering the \textit{log-sum-exp} trick for numerical stability, we change the weight of $e^{-\mathbf{z}_{\mathbf{y}_i}}$ to the bias term of $\mathbf{z}_{\mathbf{y}_i}$:
\begin{equation}
\label{eq:bal-sigmoid}
\begin{aligned}
    \hat{\sigma}(\mathbf{z}_{\mathbf{y}_i}) &= \frac{1}{1 + \frac{1-\pi_{\mathbf{y}_i}}{\pi_{\mathbf{y}_i}} \cdot e^{-\mathbf{z}_{\mathbf{y}_i}}} = \frac{1}{1 + e^{-\mathbf{z}_{\mathbf{y}_i} + \log{\frac{1-\pi_{\mathbf{y}_i}}{\pi_{\mathbf{y}_i}}}}} \\
    &= \frac{1}{1 + e^{-\mathbf{z}_{\mathbf{y}_i} + \log{(1-\pi_{\mathbf{y}_i})} - \log {\pi_{\mathbf{y}_i}}}} \\
    &= \frac{1}{1 + e^{-[\mathbf{z}_{\mathbf{y}_i} + \log {\pi_{\mathbf{y}_i}} - \log{(1-\pi_{\mathbf{y}_i})]}}}
\end{aligned}
\end{equation}
Hence, we derive the bias item of logit $\mathbf{z}_{i}$ shall be $\mathcal{B}^{\text{bce}}_{\mathbf{y}_i}=\log \pi_{\mathbf{y}_i} - \log (1-\pi_{\mathbf{y}_i})$. If we bring Eq.~\ref{eq:bal-sigmoid} into Binary CE (Eq.~\ref{eq:bce}), we will get the Balanced Binary CE as Eq.~\ref{eq:balbce}.\\ \qed

\noindent\textbf{Interpretation.} With the additional $-\log (1-\pi_{\mathbf{y}_i})$, $\mathcal{B}^{\text{bce}}_{\mathbf{y}_i}$ keeps consistent character with $\mathcal{B}^{\text{ce}}_{\mathbf{y}_i}$ \textit{w.r.t.} $\pi_{\mathbf{y}_i}$. 
Similar to $\mathcal{B}^{\text{ce}}_{\mathbf{y}_i}$, it enlarges the margins to increase the difficulty of the tail (smaller $\pi_{\mathbf{y}_i}$). 
However, $\mathcal{B}^{\text{bce}}_{\mathbf{y}_i}$ further reduces the head (larger $\pi_{\mathbf{y}_i}$) inter-class distances with larger positive values.
Notice that BCE is not class-wise mutually exclusive, and the smaller head inter-class distance helps the networks focus more on the tail's contributions.
See visualizations and more in-depth analysis in Appendix.

Through Bayesian theory \cite{PriorLT}, we can further extend the proposed Balanced BCE if the test distribution is available as $\pi^t$, which can be summarized as the following theorem:
\begin{theorem}
\label{thm:03_balance_sigmoid_test}
\normalfont{\hl{Logit Bias of Balanced BCE with Test Prior.}} Let $\pi^s_{\mathbf{y}_i}$ and $\pi^t_{\mathbf{y}_i}$ be the label $\mathbf{y}_i$ training and test distribution. If we implement the balanced cross-entropy loss via logit adjustment, the bias item of logit $\mathbf{z}_{\mathbf{y}_i}$ will be: 
\begin{equation*}
\begin{aligned}
\mathcal{B}^{\text{bce}}_{\mathbf{y}_i}=(\log \pi^s_{\mathbf{y}_i} - \log \pi^t_{\mathbf{y}_i}) - (\log (1-\pi^s_{\mathbf{y}_i}) - \log (1-\pi^t_{\mathbf{y}_i}))
\end{aligned}
\end{equation*}
\textbf{\textit{Proof.}} See detailed derivation in Appendix.
\end{theorem}
Notice that for the balanced test dataset, $\pi^t_{\mathbf{y}_i}=1/C$. Hence, the logit bias in Thm.\ref{thm:03_balance_sigmoid_test} will be:
\begin{equation}
\begin{aligned}
    \mathcal{B}^{\text{bce}}_{\mathbf{y}_i}&=(\log \pi^s_{\mathbf{y}_i} - \log 1/C) - (\log (1-\pi^s_{\mathbf{y}_i}) - \log (\frac{C-1}{C})) \\
    & = \log \pi^s_{\mathbf{y}_i} - \log (1-\pi^s_{\mathbf{y}_i}) + \log(C-1)
\end{aligned}
\label{eq:final-bias}
\end{equation}

Compared with the conclusion of Thm.~\ref{thm:02_balance_sigmoid}, we get an extra term $\log(C-1)$. From the convex objectives optimization view, there is no expected difference between Thm.~\ref{thm:02_balance_sigmoid} and Eq.~\ref{eq:final-bias}. However, it will increase ViTs' training stability remarkably, especially when the class number $C$ gets larger.

\subsection{Pipeline}
\label{sec:pipeline}
We describe \name training pipeline precisely in Alg.~\ref{alg:pipeline}, which can be divided into two stages, \textit{i.e.}, MGP and BFT. Specifically, in the MGP stage, we adopt simple data augmentation $\mathcal{A}_{pt}$ and more training epochs $T_{pt}$ to update the parameters of $\mathscr{F}$ and $\mathscr{D}$. In the BFT stage, the decoder $\mathscr{D}$ is discarded. We adopt more general data augmentations $\mathcal{A}_{ft}$ to finetune a few epochs $T_{ft}$. As shown in Alg.~\ref{alg:pipeline} Line 16, we add a hyper-parameter $\tau$ to control the influence of the proposed bias. It is worth noticing that \textit{the proposed logit bias will add negligible computational costs}. With Balanced Binary CE loss, we further optimize the parameters of $\mathscr{F}$ and $\mathscr{W}$ to achieve satisfying networks.
 
\renewcommand{\algorithmicrequire}{\textbf{Input:}}
\renewcommand{\algorithmicensure}{\textbf{Output:}}

\begin{algorithm}[t]
\caption{\name Training Pipeline.}
\label{alg:pipeline}
    \begin{algorithmic}[1]
        \Require $\mathcal{D}$, $\mathscr{F}$, $\mathscr{W}$, $\mathscr{D}$,  $T_{pt}$, $T_{ft}$, $\mathcal{A}_{pt}$, $\mathcal{A}_{ft}$, $\pi_{\mathbf{y}_i}$, $\tau$
        \Ensure Optimized $\theta_f$, $\theta_w$.
        \vspace{0.05in}
        \hrule
        \vspace{0.05in}
        \State Initialize {$\theta_f$, $\theta_d$} randomly. \Comment{MGP Stage}
        \vspace{0.05in}
        \For{$t=1$ {\bfseries to} $T_{pt}$} 
        \vspace{0.05in}
            \For {$\{\mathbf{x}, \mathbf{y}\}$ sampled from $\mathcal{D}$}
            \vspace{0.05in}
            \State $\mathbf{x} := \mathcal{A}_{pt}(\mathbf{x})$
            \vspace{0.05in}
            \State $\hat{\mathbf{x}}=\mathscr{D}\left(\mathscr{F}(\mathbf{M} \odot \mathbf{x}\ |\ \theta_f)\ |\  \theta_d \right)$ 
            \vspace{0.05in}
            \State $\mathcal{L}_{MSE}(\hat{\mathbf{x}}, \mathbf{x}) = ||\hat{\mathbf{x}}-\mathbf{x}||_2$
            \vspace{0.05in}
            \State $\{\theta_f, \theta_d\} \leftarrow \{\theta_f, \theta_d\} - \alpha \nabla_{\{\theta_f, \theta_d\}} \cdot \mathcal{L}_{MSE}(\hat{\mathbf{x}}, \mathbf{x})$
            \EndFor
        \EndFor
        \vspace{0.05in}
        \State Initialize {$\theta_w$} randomly. \Comment{BFT Stage}
        \vspace{0.05in}
        \State Calculate logit bias $\mathcal{B}^{\text{bce}}_{\mathbf{y}_i}$ via Eq.~\ref{eq:final-bias}.
        \vspace{0.05in}
        \For{$t=1$ {\bfseries to} $T_{ft}$}
        \vspace{0.05in}
            \For {$\{\mathbf{x}, \mathbf{y}\}$ sampled from $\mathcal{D}$}
            \vspace{0.05in}
            \State $\mathbf{x} := \mathcal{A}_{ft}(\mathbf{x})$
            \vspace{0.05in}
            \State $\mathbf{v} = \mathscr{F}(\mathbf{x}\ |\ \theta_f)$ 
            \vspace{0.05in}
            \State $\mathbf{z} = \mathscr{W}(\mathbf{v}\ |\ \theta_w) + \tau \cdot \mathcal{B}^{\text{bce}}$ 
            \vspace{0.05in}
            \State Calculate $\mathcal{L}_{BCE}$ via Eq.~\ref{eq:bce} with calibrated $\mathbf{z}$.
            \vspace{0.05in}
            \State $\{\theta_f, \theta_w\} \leftarrow \{\theta_f, \theta_w\} - \alpha \nabla_{\{\theta_f, \theta_w\}} \cdot \mathcal{L}_{BCE}$
            \EndFor
        \vspace{0.05in}
        \EndFor
    \end{algorithmic}
\end{algorithm}

\setlength{\textfloatsep}{5pt}

\section{Experiment}

\subsection{Datasets}
\label{sec:dataset}
\textbf{CIFAR-10/100-LT} are created from the original CIFAR datasets \cite{Cifar}, where $\gamma$ controls the data imbalance degree. Following previous works \cite{BBN, LDAM, PriorLT, PaCo}, we employ imbalance factors \{100, 10\} in our experiments. \textbf{ImageNet-LT/BAL} are both the subsets of popular ImageNet \cite{Imagenet}. The \textit{LT} version \cite{OLTR} ($\gamma=256$) is selected following the \textit{Pareto} distribution with power value $\alpha=6$, which contains 115.8K images from 1,000 categories. We build the \textit{BAL} version ($\gamma=1$) by sampling 116 images per category to exploit how ViTs perform given a similar number of training images. Notice that both LT and BAL adopt the \textit{same} validation dataset. \textbf{iNaturalist 2018} \cite{iNat, iNatFinegrained} (iNat18 for short) is a species classification dataset, which contains 437.5K images from 8,142 categories and suffers from extremely LTR problem ($\gamma=512$). \textbf{Places-LT} is a synthetic long-tail variant of the large-scale scene classification dataset Places \cite{PlaceLT}. With 62.5K images from 365 categories, its class cardinality ranges from 5 to 4,980 ($\gamma=996$). All datasets adopt the official validation images for fair comparisons. See detailed dataset information in Appendix.

\begin{table}[t]
\centering
\caption{Top-1 accuracy (\%) of ResNet50 on ImageNet-LT. $\dagger$ indicates results with ResNeXt50. $*$: training with 384 resolution.}
\resizebox{\linewidth}{53mm}{
\begin{tabular}{l|c|ccc|c}
\toprule
\multicolumn{1}{l|}{Method} & Ref.      & Many & Med.        & Few           & Acc  \\ \midrule
CE \cite{CB}                            & CVPR 19            & 64.0 & 33.8   & 5.8  & 41.6 \\
LDAM \cite{LDAM}                        & NeurIPS 19         & 60.4 & 46.9   & 30.7 & 49.8 \\
c-RT \cite{NCM}                         & ICLR 20            & 61.8 & 46.2   & 27.3 & 49.6 \\
$\tau$-Norm \cite{NCM}                  & ICLR 20            & 59.1 & 46.9   & 30.7 & 49.4 \\
Causal \cite{CausalNorm}                & NeurIPS 20         & 62.7 & 48.8   & 31.6 & 51.8 \\
Logit Adj. \cite{LA}                    & ICLR 21            & 61.1 & 47.5   & 27.6 & 50.1 \\
RIDE(4E)$\dagger$ \cite{RIDE}           & ICLR 21            & 68.3 & 53.5   & 35.9 & 56.8 \\
MiSLAS \cite{MiSLAS}	                & CVPR 21	         & 62.9	& 50.7	 & 34.3	& 52.7 \\
DisAlign \cite{DisAlign}                & CVPR 21            & 61.3 & 52.2   & 31.4 & 52.9 \\
ACE$\dagger$ \cite{ACE}                 & ICCV 21            & 71.7 & 54.6   & 23.5 & 56.6 \\
PaCo$\dagger$ \cite{PaCo}               & ICCV 21            & 68.0 & 56.4   & 37.2 & 58.2 \\
TADE$\dagger$ \cite{TADE}               & ICCV 21            & 66.5 & 57.0   & \textbf{43.5} & 58.8 \\
TSC \cite{TSC}                          & CVPR 22            & 63.5 & 49.7   & 30.4 & 52.4 \\
GCL \cite{GCL}                          & CVPR 22            & 63.0 & 52.7   & 37.1 & 54.5 \\
TLC \cite{TLC}                          & CVPR 22            & 68.9 & 55.7   & 40.8 & 55.1 \\
BCL$\dagger$ \cite{BCL}                 & CVPR 22            & 67.6 & 54.6   & 36.6 & 57.2 \\
NCL \cite{NCL}                          & CVPR 22            & 67.3 & 55.4   & 39.0 & 57.7 \\
SAFA \cite{SAFA}                        & ECCV 22            & 63.8 & 49.9   & 33.4 & 53.1 \\
DOC \cite{DOC}                          & ECCV 22            & 65.1 & 52.8   & 34.2 & 55.0 \\
DLSA \cite{DLSA}                        & ECCV 22            & 67.8 & 54.5   & 38.8 & 57.5 \\ \hline
\multicolumn{6}{c}{\cellcolor{lightgrey}ViT-B training from scratch} \\
ViT \cite{ViT}                          & ICLR 21           & 50.5 & 23.5   & 6.9 & 31.6 \\
MAE \cite{MAE}                          & CVPR 22           & 74.7 & 48.2   & 19.4 & 54.5 \\
DeiT \cite{deit3}                       & ECCV 22           & 70.4     & 40.9     & 12.8     & 48.4 \\
\name                        & \multirow{2}{*}{-} & 73.6 & 56.4   & 41.0 & 60.9 \\
\name $^*$                   &                    & \textbf{76.4} & \textbf{59.7}   & 42.7 & \textbf{63.8} \\ 
\bottomrule
\end{tabular}}
\label{tab:performance-imagenet-lt}
\end{table}

\subsection{Implement Details}
For image classification on main benchmarks, we adopt ViT-Base-16 \cite{ViT} as the backbone and ViT-Tiny / Small \cite{deit3} ViT-Large \cite{ViT} for the ablation study. All models are trained with AdamW optimizer \cite{AdamW} with $\beta s= \{0.9, 0.95\}$. The effective batch size is 4,096 (MGP) / 1,024 (BFT). Vanilla ViTs\cite{ViT}, DeiT III\cite{deit3} and MAE\cite{MAE} are all trained 800 epochs because ViTs require longer training time to converge. Following previous work\cite{MAE}, \name is pretrained 800 epochs with the mask ratio 0.75 and finetuned 100(50) epochs for ViT-T/S/B(L). We train all models with RandAug(9, 0.5) \cite{Randaugment}, mixup (0.8) and cutmix (1.0). All experiments set $\tau\equiv1$. For fair comparisons, we re-implement \cite{CB, LDAM, LADE, IB, BS} with ViTs in the same settings. Following \cite{OLTR}, we report Top-1 accuracy and three groups' accuracy: Many-shot ($>$100 images), Medium-shot (20$\sim$100 images) and Few-shot ($<$20 images). Besides, we report the Expected Calibration Error (ECE) and Maximum Calibration Error (MCE) to quantify the predictive uncertainty \cite{ECE}. See detailed implementation settings in Appendix.

\subsection{Comparison with Prior Arts}

We conduct comprehensive experiments with ViT-B-16 on ImageNet-LT, iNat18, and Place-LT benchmarks. \name successfully trains it \textbf{from scratch} without any additional data pretraining and outperforms ResNet50, ResNeXt50 and ResNet152 conspicuously.

\begin{table}[t]
\centering
\caption{Top-1 accuracy (\%) of ResNet50 on iNaturalist 2018. $*$: training with 384 resolution.}
\resizebox{\linewidth}{62mm}{
\begin{tabular}{@{}l|c|ccc|c@{}}
\toprule
Method   & Ref.         & Many & Med. & Few  & Acc  \\ \midrule
CE \cite{CB}            & CVPR 19 & 72.2 & 63.0 & 57.2 & 61.7 \\
OLTR \cite{OLTR}        & CVPR 19 & 59.0 & 64.1 & 64.9 & 63.9 \\
c-RT \cite{NCM}         & ICLR 20 & 69.0 & 66.0 & 63.2 & 65.2 \\
$\tau$-Norm \cite{NCM}  & ICLR 20 & 65.6 & 65.3 & 65.9 & 65.6 \\
LWS \cite{NCM}          & ICLR 20 & 65.0 & 66.3 & 65.5 & 65.9 \\
BBN \cite{BBN}          & CVPR 20 & 61.8 & 73.6 & 66.9 & 69.6 \\
BS \cite{BS}            & ICLR 21 & 70.0 & 70.2 & 69.9 & 70.0 \\
RIDE(4E) \cite{RIDE}    & ICLR 21 & 70.9 & 72.5 & 73.1 & 72.6 \\
DisAlign \cite{DisAlign}& CVPR 21 & 69.0 & 71.1 & 70.2 & 70.6 \\
MiSLAS \cite{MiSLAS}    & CVPR 21 & 73.2 & 72.4 & 70.4 & 71.6 \\
DiVE \cite{DiVE}        & ICCV 21 & 70.6 & 70.0 & 67.6 & 69.1 \\
ACE(4E) \cite{ACE}      & ICCV 21 & -    & -    & -    & 72.9 \\
TADE \cite{TADE}        & ICCV 21 & 74.4 & 72.5 & 73.1 & 72.9 \\
PaCo \cite{PaCo}        & ICCV 21 & 70.4 & 72.8 & 73.6 & 73.2 \\
ALA \cite{ALA}          & AAAI 22 & 71.3 & 70.8 & 70.4 & 70.7 \\
TSC \cite{TSC}          & CVPR 22 & 72.6 & 70.6 & 67.8 & 69.7 \\
LTR-WD \cite{LTR-WD}    & CVPR 22 & 71.2 & 70.4 & 69.7 & 70.2 \\
GCL \cite{GCL}          & CVPR 22 & 67.5 & 71.3 & 71.5 & 71.0 \\
BCL \cite{BCL}          & CVPR 22 & 66.7 & 71.0 & 70.7 & 70.4 \\
NCL \cite{NCL}          & CVPR 22 & 72.0 & 74.9 & 73.8 & 74.2 \\
DOC \cite{DOC}          & ECCV 22 & 72.8 & 71.7 & 70.0 & 71.0 \\
DLSA \cite{DLSA}        & ECCV 22 & -    & -    & -    & 72.8 \\ \hline
\multicolumn{6}{c}{\cellcolor{lightgrey}ViT-B training from scratch} \\
ViT \cite{ViT}          & ICLR 21 & 65.4 & 55.3 & 50.9 & 54.6 \\
MAE \cite{MAE}          & CVPR 22 & 79.6 & 70.8 & 65.0 & 69.4 \\
DeiT \cite{deit3}       & ECCV 22 & 72.9 & 62.8 & 55.8 & 61.0 \\
\name                   & -       & 78.9 & 76.5 & 74.8 & 76.1 \\
\name $^*$              & -       & \textbf{83.2} & \textbf{81.5} & \textbf{79.7} & \textbf{81.0} \\ \bottomrule
\end{tabular}}
\label{tab:performance-iNat}
\end{table}

\begin{table}[t]
\centering
\caption{Top-1 accuracy (\%) of ResNet152 (with ImageNet-1K pretrained weight) on Places-LT. $*$: training with 384 resolution.}
\resizebox{\linewidth}{52mm}{
\begin{tabular}{@{}l|c|ccc|c@{}}
\toprule
Method    & Ref.       & Many & Med. & Few  & Acc  \\ \midrule
CE \cite{CB}       & CVPR 19    & 45.7 & 27.3 & 8.2  & 30.2 \\
Focal \cite{Focal}    & ICCV 17    & 41.1 & 34.8 & 22.4 & 34.6 \\
Range \cite{Range}    & CVPR 17    & 41.1 & 35.4 & 23.2 & 35.1 \\
OLTR \cite{OLTR}     & CVPR 19    & 44.7 & 37.0 & 25.3 & 35.9 \\
FSA \cite{FSA}      & ECCV 20    & 42.8 & 37.5 & 22.7 & 36.4 \\
LWS \cite{NCM}      & ICLR 20    & 40.6 & 39.1 & 28.6 & 37.6 \\
Causal \cite{CausalNorm}   & NeurIPS 20 & 23.8 & 35.8 & \textbf{40.4} & 32.4 \\
BS \cite{BS}       & NeurIPS 20 & 42.0 & 39.3 & 30.5 & 38.6 \\
DisAlign \cite{DisAlign} & CVPR 21    & 40.4 & 42.4 & 30.1 & 39.3 \\
LADE \cite{LADE}     & CVPR 21    & 42.8 & 39.0 & 31.2 & 38.8 \\
RSG \cite{RSG}     & CVPR 21    & 41.9 & 41.4 & 32.0 & 39.3 \\
TADE \cite{TADE}     & ICCV 21    & 43.1 & 42.4 & 33.2 & 40.9 \\
PaCo \cite{PaCo}     & ICCV 21    & 36.1 & \textbf{47.9} & 35.3 & 41.2 \\
ALA \cite{ALA}      & AAAI 22    & 43.9 & 40.1 & 32.9 & 40.1 \\
NCL \cite{NCL}      & CVPR 22    & -    & -    & -    & 41.8 \\
BF \cite{BatchFormer} & CVPR 22    & 44.0 & 43.1 & 33.7 & 41.6 \\
CKT \cite{CKT}      & CVPR 22    & 41.6 & 41.4 & 35.1 & 40.2 \\
GCL \cite{GCL}      & CVPR 22    & -    & -    & -    & 40.6 \\ 
Bread \cite{Bread}  & ECCV 22    & 40.6 & 41.0 & 33.4 & 39.3 \\ \hline
\multicolumn{6}{c}{\cellcolor{lightgrey}ViT-B training from scratch} \\
MAE \cite{MAE}          & CVPR 22 & 48.9 & 24.6 & 8.7 & 30.3 \\
DeiT \cite{deit3}       & ECCV 22 & \textbf{51.6} & 31.0 & 9.4 & 34.2 \\
\name       & -     & 48.1    & 40.6    & 27.5    & 40.8    \\
\name $^*$  & -     & 50.7    & 42.4    & 27.9    & \textbf{42.6}    \\ \bottomrule
\end{tabular}}
\vspace{-10pt}
\label{tab:performance-place}
\end{table}
\begin{table}[t]
\centering
\caption{The transfer performance of ViT-B (resolution 224$\times$224) on iNat18 dataset. D-PT represents the pretrain datasets. BAL and LT have similar amounts of data and contribute to similar transfer performance, which means MGP is robust to data distribution.}
\resizebox{\linewidth}{13mm}{
\begin{tabular}{@{}c|c|ccc|c|cc@{}}
\toprule
D-PT    & Loss    & Many  & Med.  & Few   & Acc   & ECE  & MCE  \\ \midrule
BAL     & CE      & 63.7 & 57.1 & 52.4  & 55.9 & 1.2 & 3.4 \\
LT      & CE      & 64.5 & 57.5 & 52.7 & 56.4 & 1.2 & 3.1 \\ \midrule
BAL     & Bal-BCE & 53.3 & 58.8  & 60.7 & 59.0 & 0.8 & 1.6 \\
LT      & Bal-BCE & 56.5 & 60.8 & 61.6 & 60.7 & 1.0 & 2.9 \\ \bottomrule
\end{tabular}}
\label{tab:performance-transfer}
\end{table}

\noindent\textit{\textbf{Comparison on ImageNet-LT.}} Tab.~\ref{tab:performance-imagenet-lt} shows the experimental comparison results with recent SOTA methods on ImageNet-LT. The training resolution of \name is 224 / 224 for MGP / BFT. Based on the model ensemble, multi-expert methods like RIDE\cite{RIDE}, TADE\cite{TADE}, and NCL \cite{NCL} exhibit powerful preference with heavier model size compared to baseline. The CL-based methods (PaCo\cite{PaCo}, TSC \cite{TSC}, BCL\cite{BCL}) also achieve satisfying results with larger batches and longer training epochs. However, our \name has shown superior performance without bells and whistles and outperforms them consistently on all metrics while training ViTs from scratch. Notice that \name gains more performance (63.8\% vs 60.9\%) with higher image resolution in the BFT stage, which is consistent with the observations in \cite{FixRes, Swin, deit3}. Notice that \name improves the iNat18 dataset most significantly because BCE mitigates fine-grained problems as well \cite{iNatFG}. 

\begin{table*}[t]
\centering
\caption{Ablation study of the proposed bias (\textit{c.f.} Eq.~\ref{eq:final-bias}) on CE / BCE. All models are trained on ImageNet-LT with the same settings. Our Bal-BCE ameliorates the original BCE by a large margin in all aspects, which is consistent with CE and Bal-CE.}
\resizebox{0.85\linewidth}{40mm}{
\begin{tabular}{@{}c|c|c|lll|l|ll@{}}
\toprule
Model                  & Size                  & Loss    & \multicolumn{1}{l}{Many $\uparrow$} & \multicolumn{1}{l}{Med. $\uparrow$} & \multicolumn{1}{l|}{Few $\uparrow$} & \multicolumn{1}{l|}{Acc $\uparrow$} & \multicolumn{1}{l}{ECE $\downarrow$} & \multicolumn{1}{l}{MCE $\downarrow$} \\ \midrule
\multirow{4}{*}{ViT-Tiny \cite{deit3}}  & \multirow{4}{*}{5.7M} & CE      & 56.1                     & 29.2                       & 10.5                     & 37.0                     & 3.7                     & 6.1                     \\
                       &                       & Bal-CE  & 48.8 \bad{(-7.3)}        & 39.2 \good{(+10.0)}        & 28.1 \good{(+17.6)}      & \textbf{41.4} \good{(+4.4) }      & 2.6 \good{(-1.1)}       & 4.6 \good{(-1.6)}       \\ \cmidrule(l){3-9} 
                       &                       & BCE     & 42.1                     & 11.1                       & 0.9                      & 21.6                     & 2.9                     & 8.6                     \\
                       &                       & Bal-BCE & 50.6 \good{(+8.4)}       & 37.2 \good{(+26.1)}        & 26.1 \good{(+25.2)}      & 40.8 \good{(+19.2)}      & 3.1 \bad{(+0.1)}        & 6.8 \good{(-1.8)}       \\ \midrule
\multirow{4}{*}{ViT-Small \cite{deit3}} & \multirow{4}{*}{22M}  & CE      & 68.9                     & 43.1                       & 17.3                     & 49.5                     & 4.7                     & 9.2                     \\
                       &                       & Bal-CE  & 62.7 \bad{(-6.2)}        & 52.0 \good{(+8.9)}         & 36.3 \good{(+19.0)}      & 54.0 \good{(+4.5)}       & 0.9 \good{(-3.8)}       & 2.4 \good{(-6.8)}       \\ \cmidrule(l){3-9} 
                       &                       & BCE     & 62.4                     & 30.6                       & 8.4                      & 39.8                     & 5.7                     & 11.1                    \\
                       &                       & Bal-BCE & 65.8 \good{(+3.4)}       & 50.6 \good{(+20.0)}        & 32.9 \good{(+24.6)}      & \textbf{54.1} \good{(+14.2)}      & 4.8 \good{(-0.9)}       & 9.0 \good{(-2.2)}       \\ \midrule
\multirow{4}{*}{ViT-Base \cite{ViT}}  & \multirow{4}{*}{86M}  & CE      & 74.7                     & 48.2                       & 19.4                     & 54.5                     & 5.1                     & 6.8                     \\
                       &                       & Bal-CE  & 70.5 \bad{(-4.3)}        & 56.8 \good{(+8.6)}         & 43.7 \good{(+24.3)}      & 60.1 \good{(+5.6)}       & 3.7 \good{(-1.4)}       & 4.9 \good{(-1.9)}       \\ \cmidrule(l){3-9} 
                       &                       & BCE     & 73.7                     & 46.5                       & 15.6                     & 52.4                     & 5.6                     & 7.9                     \\
                       &                       & Bal-BCE & 73.6 \bad{(-0.1)}        & 55.8 \good{(+9.3)}         & 41.0 \good{(+25.4)}      & \textbf{60.9} \good{(+8.6)}       & 2.4 \good{(-3.1)}       & 3.2 \good{(-4.7)}       \\ \midrule
\multirow{4}{*}{ViT-Large \cite{ViT}} & \multirow{4}{*}{304M} & CE      & 77.3                     & 51.5                       & 21.7                     & 57.4                     & 3.6                     & 7.4                     \\
                       &                       & Bal-CE  & 72.7 \bad{(-4.5)}        & 60.1 \good{(+8.6)}         & 41.9 \good{(+20.3)}      & 62.1 \good{(+4.8)}       & 2.1 \good{(-1.5)}       & 4.2 \good{(-3.2)}       \\ \cmidrule(l){3-9} 
                       &                       & BCE     & 74.7                     & 46.7                       & 17.0                     & 53.4                     & 8.4                     & 15.9                    \\
                       &                       & Bal-BCE & 75.3 \good{(+0.6)}       & 58.8 \good{(+12.1)}        & 37.5 \good{(+20.5)}      & \textbf{62.6} \good{(+9.2)}       & 6.6 \good{(-1.8)}       & 14.8 \good{(-1.1)}      \\ \bottomrule
\end{tabular}}
\vspace{-10pt}
\label{tab:bias-ablation}
\end{table*}

\noindent\textit{\textbf{Comparison on iNaturalist 2018.}} Tab.~\ref{tab:performance-iNat} lists experimental results on iNaturalist 2018. The training resolution of \name is 128 / 224 for MGP / BFT. \name consistently surpasses recent SOTA methods like PcCo\cite{PaCo}, NCL\cite{NCL} and DLSA\cite{DLSA}. Unlike most LTR methods, our \name improves all groups' Acc without sacrificing many-shot performance. Compared to ensemble NCL (3$\times$), \name surpasses it by 1.9\% (6.8\% higher resolution) with comparable model size, which verifies the effectiveness of \name.

\noindent\textit{\textbf{Comparison on Places-LT.}}
Tab.~\ref{tab:performance-place} summarizes the experimental results on Places-LT. All LTR proposals adopt ResNet152 pre-trained on ImageNet-1K. For fair comparisons, we conduct MGP at ImageNet-1K and BFT at Places-LT. As illustrated in Tab.~\ref{tab:performance-place}, \name obtains satisfying performance compared with previous SOTAs. Notice that Places-LT has limited instances compared to iNat18 (437.5K) and ImageNet-1K (1M). Considering both Tab.~\ref{tab:performance-iNat} and Tab.~\ref{tab:performance-place} results, we observe that ViTs, which benefit from large-scale data, are limited in this case. However, our \name performs the best even in such data paucity situations.

\subsection{Further Analysis}
\noindent\textit{\textbf{Robustness of MGP.}}
The performance results in Tab.~\ref{tab:pretrain-recips} have shown that MGP is more robust to learning label irrelevant features than supervised methods. For deeper observations, we show the transfer results in Tab.~\ref{tab:performance-transfer}. Concretely, we conduct MGP on ImageNet-LT / ImageNet-BAL (See section~\ref{sec:dataset}) and BFT on iNat18 with resolution 224. Regardless of the data distribution of the MGP dataset, both BAL and LT achieve quite similar performance in terms of all evaluated metrics on iNat18. If we further compare the reported results with Tab.~\ref{tab:performance-iNat}, we will draw the conclusion that the training instance number plays the key role in \name instead of the label distribution, which is clearly different from previous SCL \cite{PaCo, BCL} methods. We show more  reconstruction visualization given by LT / BAL in Appendix. 

\begin{figure}[t]
\centering
\includegraphics[width=\linewidth]{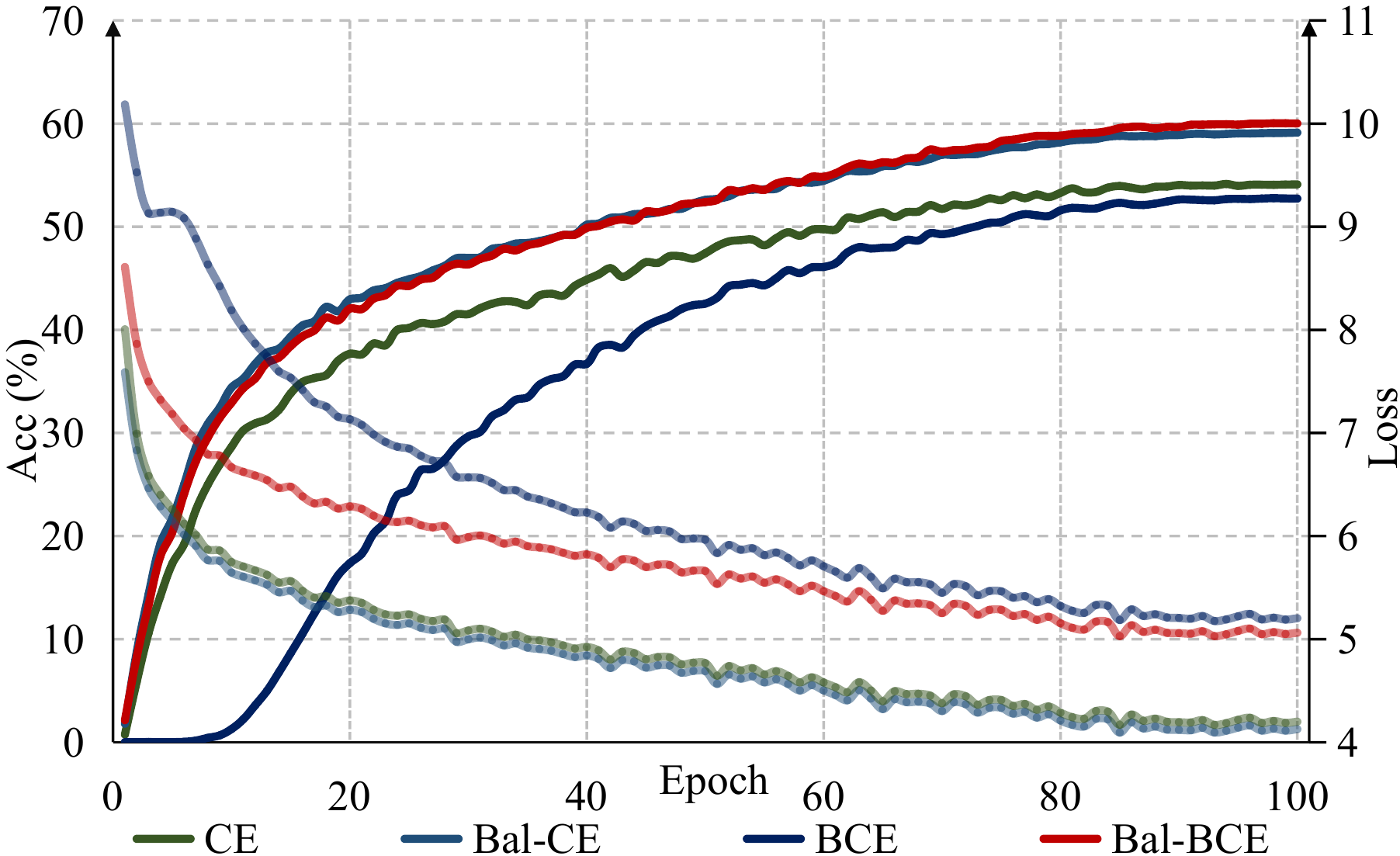}
\caption{Training loss and Top-1\% accuracy of ViT-S on iNaturalist 2018 dataset. Solid and dot lines represent the accuracy and training loss, respectively. All models adopt the same settings and random seed except for loss type.}
\label{fig:loss-acc}
\end{figure}

\noindent\textit{\textbf{Effectiveness of Proposed Bias.}}
To learn balanced ViTs, we propose Bal-BCE with a simple yet effective logit bias (\textit{c.f.} Eq.~\ref{eq:final-bias}). To validate its effectiveness, we conduct the ablation study and compare it with the most popular re-balance loss, \textit{i.e.}, Bal-CE. As shown in Tab.~\ref{tab:bias-ablation}, the new logit bias boosts vanilla BCE significantly with lower ECE on four ViT backbones, which is consistent with the behavior of Bal-CE. It is worth noticing that CE generally performs better than BCE in LTR scenarios, which is different from the conclusion in balanced datasets \cite{deit3}. However, our Bal-BCE alleviates it remarkably and outperforms Bal-CE in most cases. In addition, Bal-BCE shows more satisfying numerical stability and faster convergence. See Bal-BCE in Fig.~\ref{fig:loss-acc}. for detailed illustrations.

For comprehensive comparisons, we re-implement recent rebalancing strategies in our BFT stage and show the results of ViT-B on CIFAR-LT in Tab.~\ref{tab:loss-on-vit}. Without loss of fairness, we conduct MGP on ImageNet-1K because the resolution (32$\times$32) of CIFAR is too small to mask for ViT-B-16. We do not reproduce the CL-based (conflict to MGP) and ensemble (memory limitation) methods. We also give up some ingenious rebalancing methods for loss NaN during training. As shown in Tab.~\ref{tab:loss-on-vit}, the proposed Bal-BCE achieves the best results, which firmly manifests its effectiveness. Notice that some methods are not consistent with their performance on ResNet, which means some exquisite designs may not generalize well on ViTs.

\begin{figure}[t]
	\flushleft
	\subfloat[CIFAR-100-LT-10]{
        \includegraphics[width=0.98\linewidth]{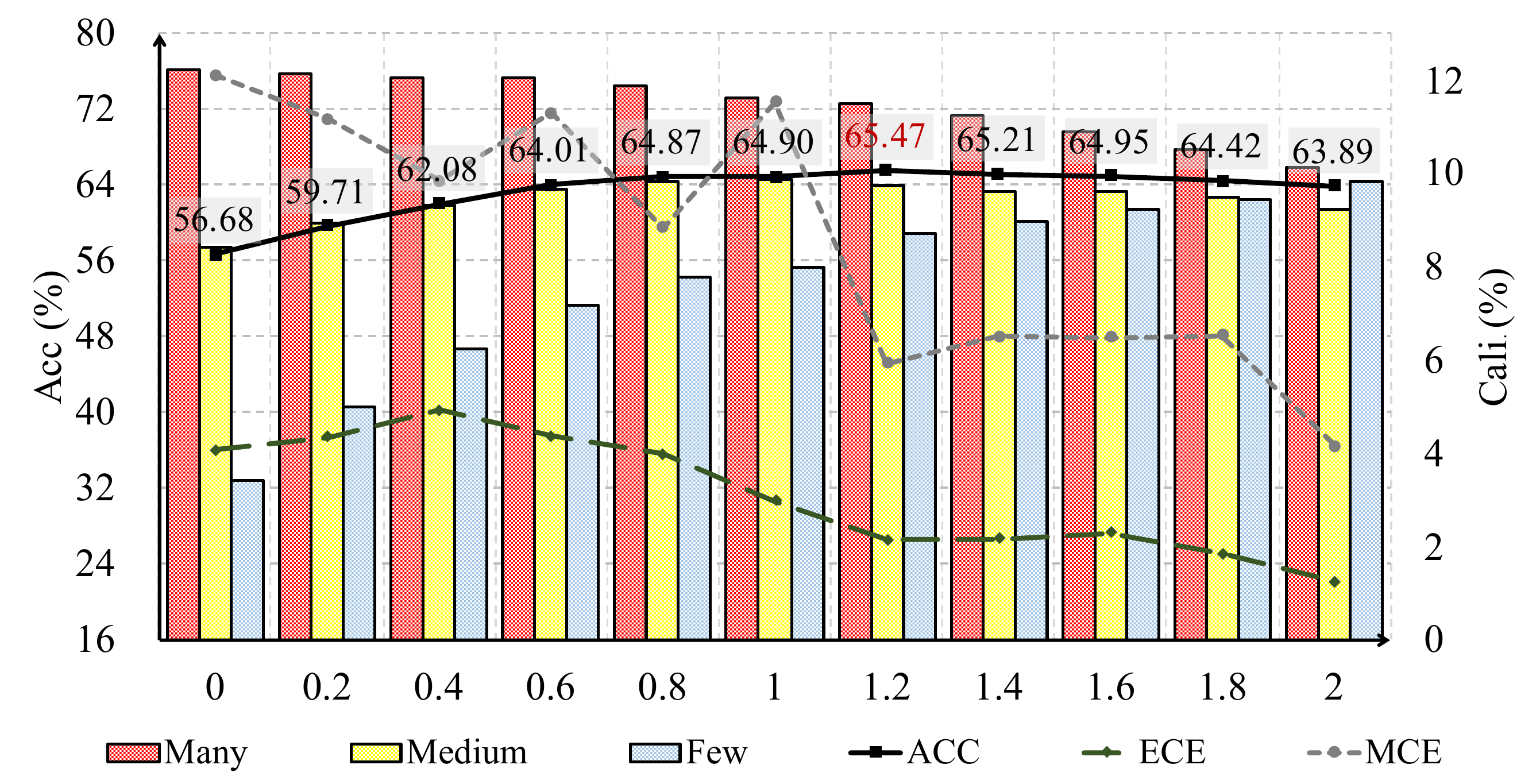}
        \label{fig:cifar100-10}
    }
    \\
    \subfloat[CIFAR-100-LT-100]{
        \includegraphics[width=0.98\linewidth]{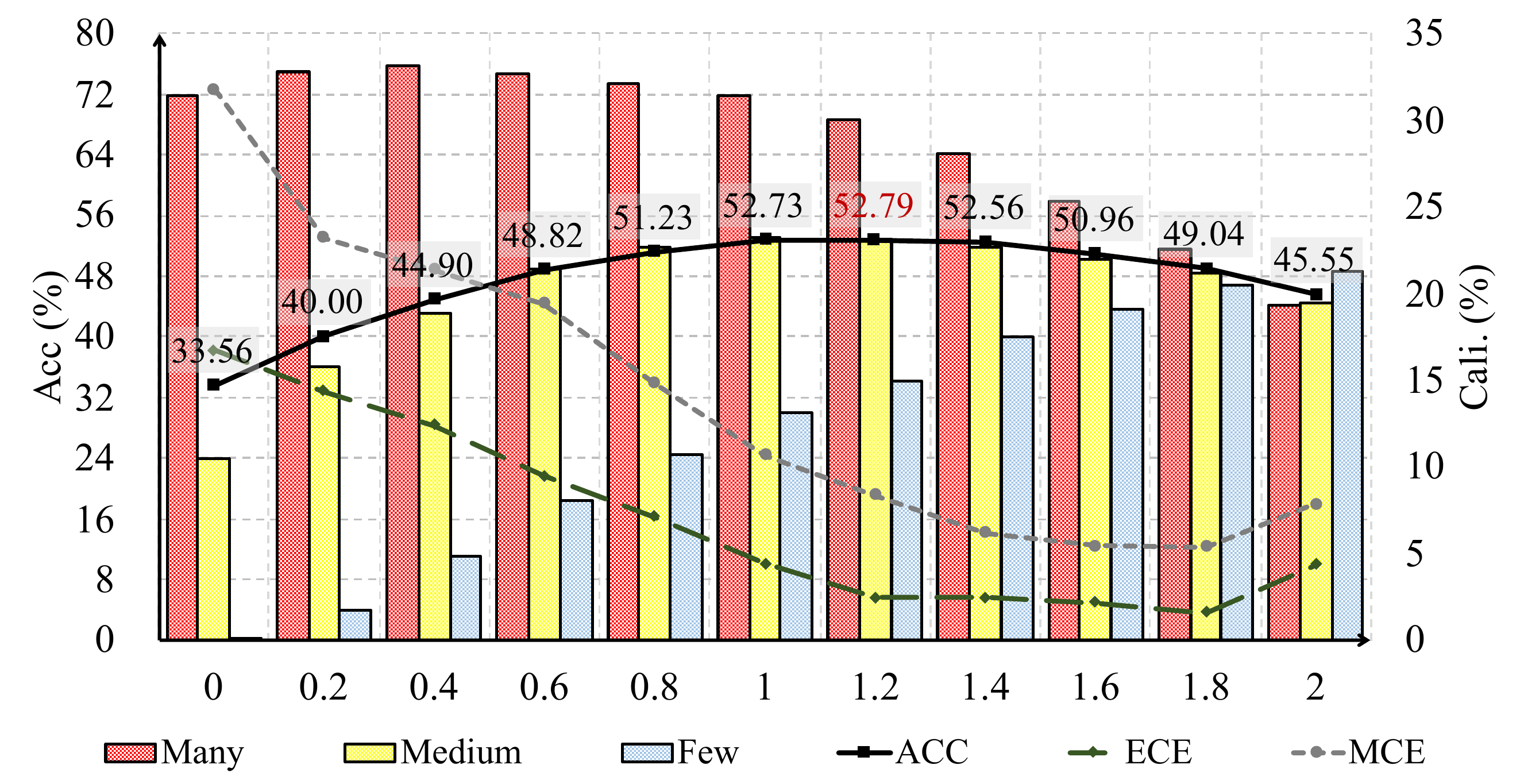}
        \label{fig:cifar100-100}
    }
	\caption{Performance of ViT-B with different $\tau$ on CIFAR-100-LT. A bigger $\tau$ results in better few-shot performance.}
\label{fig:hyp-tau} 
\end{figure}

\noindent\textit{\textbf{Hyper-Parameter Analysis.}}
In Alg.~\ref{alg:pipeline} Line 11, we add a hyper-parameter $\tau$ to adjust our proposed bias (Eq.~\ref{eq:final-bias}). We further present in-depth investigations on the influence of $\tau$. Similar to the aforementioned settings with plain augmentations, we conduct the ablation study on CIFAR-100-LT with MGP on ImageNet-1K and show the results in Fig.\ref{fig:hyp-tau}. The few-shot accuracy gets obvious amelioration when $\tau$ gets larger, which is consistent with our explanations in section \ref{sec:finetune}. The best overall accuracy is obtained around $1$, which inspires us to set $\tau\equiv1$ in \name for all experiments by default. Besides, the ECE gets smaller with increasing $\tau$, which means that the proposed bias guides ViTs to be the calibrated models with Fisher Consistency ensured \cite{LA}.

\begin{table}[t]
	\centering
	\caption{Ablation study of rebalancing strategies on ViT-B.}
	\setlength{\tabcolsep}{2mm}
	\begin{tabular}{@{}l|cc|cc@{}}
	\toprule
	\multicolumn{1}{c|}{Method} & \multicolumn{2}{c|}{CIFAR-10-LT} & \multicolumn{2}{c}{CIFAR-100-LT} \\ \midrule
	\multicolumn{1}{c|}{$\gamma$} & 100         & 10          & 100         & 10           \\ \midrule
	CE\cite{CB}                  & 79.2        & 89.5        & 50.9        & 66.1            \\
	CB \cite{CB}                 & 82.0        & 89.9        & 52.0        & 66.8            \\
	LDAM \cite{LDAM}             & 78.6        & 88.6        & 52.56       & 66.1            \\
	LADE \cite{LADE}             & 68.8        & 81.7        & 56.7        & 68.2            \\
	IB \cite{IB}                 & 75.4        & 79.2        & 50.8        & 51.6            \\
	Bal-CE \cite{BS}             & 84.4        & 90.7        & 56.8        & 68.1            \\
	Bal-BCE (ours)               & \textbf{86.3}  & \textbf{91.3}  & \textbf{58.2}  & \textbf{69.2}\\ \bottomrule
	\end{tabular}
\label{tab:loss-on-vit}
\end{table}

\section{Discussion}
\noindent\textit{\textbf{Why train from scratch?}} 
Previous ViTs papers are all based on pretrained weights from ImageNet-1K or ImageNet-22K and thus may lead to unfair comparisons with LTR methods, which are all trained from scratch. It is difficult to conclude that the intriguing performance mainly benefits from their proposals. Our approach provides a strong baseline to verify proposals' effectiveness with ViTs. It's also instructive to train plain ViTs for areas where data exhibits severe domain gaps. From the original intention of the LTR task, the core is to learn more large-scale imbalanced data effectively. Our work provides a feasible way to utilize more real-world LT (labels or attributes) data without expensive artificial balancing to achieve better representation learning.

\noindent\textit{\textbf{How we extend MAE.}} 
We empirically prove that masked autoencoder learns generalized features even with imbalanced data, which is quite different from other self-supervised manners like CL \cite{SimCLR} and SCL \cite{SCL}. Extensive experiments on ImageNet-LT/BAL show that \textit{the instance number is more crucial than balanced annotation}. We further propose the balanced binary cross-entropy loss to build our \name and achieve a new SOTA in LTR.

\noindent\textit{\textbf{Limitations.}}
One limitation is that \name can not be deployed in an end-to-end manner. An intuitive idea is two branches learning to optimize the decoder and classifier simultaneously, like BBN\cite{BBN} or PaCo\cite{PaCo}. However, the heavily masked image prevents effective classification, while dynamic mask ratios exacerbate memory limitations.

\section{Conclusion}
In this paper, we propose to Learn imbalanced data with Vision Transformers (LiVT), which consists of Masked Generative Pretraining (MGP) and Balanced Fine Tuning (BFT). MGP is based on our empirical insight that it guides ViTs to learn more generalized features on long-tailed datasets compared to supervised or contrastive paradigms. BFT is based on the theoretical analysis of Binary Cross-Entropy (BCE) in the imbalanced scenario. We propose the balanced BCE to learn unbiased ViTs by compensating extra logit margins. Bal-BCE ameliorates BCE significantly and surpasses the powerful and widely adopted Balanced Cross-Entropy loss when cooperating with ViTs. Extensive experiments on large-scale datasets demonstrate that \name successfully trains ViTs without any additional data and achieves a new state-of-the-art for long-tail recognition.
\clearpage

\section*{Acknowledgement}
\noindent This work was supported by the National Key R\&D Program of China (2022YFB4701400/4701402), SZSTC Grant\\(JCYJ 20190809172201639, WDZC20200820200655001), Shenzhen Key Laboratory (ZDSYS20210623092001004). 

\balance{
    {\small
    \bibliographystyle{cvpr}
    \bibliography{main}

\begin{thebibliography}{10}\itemsep=-1pt

\bibitem{LTR-WD}
Shaden Alshammari, Yu-Xiong Wang, Deva Ramanan, and Shu Kong.
\newblock Long-tailed recognition via weight balancing.
\newblock In {\em CVPR}, pages 6897--6907, 2022.

\bibitem{BeiT}
Hangbo Bao, Li Dong, Songhao Piao, and Furu Wei.
\newblock {BE}it: {BERT} pre-training of image transformers.
\newblock In {\em ICLR}, 2022.

\bibitem{ACE}
Jiarui Cai, Yizhou Wang, Jenq-Neng Hwang, et~al.
\newblock Ace: Ally complementary experts for solving long-tailed recognition
  in one-shot.
\newblock In {\em ICCV}, pages 112--121, 2021.

\bibitem{LDAM}
Kaidi Cao, Colin Wei, Adrien Gaidon, Nikos Arechiga, and Tengyu Ma.
\newblock Learning imbalanced datasets with label-distribution-aware margin
  loss.
\newblock {\em NeurIPS}, 32, 2019.

\bibitem{DETR}
Nicolas Carion, Francisco Massa, Gabriel Synnaeve, Nicolas Usunier, Alexander
  Kirillov, and Sergey Zagoruyko.
\newblock End-to-end object detection with transformers.
\newblock In {\em ECCV}, pages 213--229. Springer, 2020.

\bibitem{MaskGIT}
Huiwen Chang, Han Zhang, Lu Jiang, Ce Liu, and William~T. Freeman.
\newblock Maskgit: Masked generative image transformer.
\newblock In {\em CVPR}, June 2022.

\bibitem{SimCLR}
Ting Chen, Simon Kornblith, Mohammad Norouzi, and Geoffrey Hinton.
\newblock A simple framework for contrastive learning of visual
  representations.
\newblock In {\em ICML}, pages 1597--1607. PMLR, 2020.

\bibitem{MoCo3}
Xinlei Chen, Saining Xie, and Kaiming He.
\newblock An empirical study of training self-supervised vision transformers.
\newblock In {\em ICCV}, pages 9640--9649, 2021.

\bibitem{Remix}
Hsin-Ping Chou, Shih-Chieh Chang, Jia-Yu Pan, Wei Wei, and Da-Cheng Juan.
\newblock Remix: rebalanced mixup.
\newblock In {\em ECCV}, pages 95--110. Springer, 2020.

\bibitem{FSA}
Peng Chu, Xiao Bian, Shaopeng Liu, and Haibin Ling.
\newblock Feature space augmentation for long-tailed data.
\newblock In {\em ECCV}, pages 694--710. Springer, 2020.

\bibitem{Randaugment}
Ekin~D Cubuk, Barret Zoph, Jonathon Shlens, and Quoc~V Le.
\newblock Randaugment: Practical automated data augmentation with a reduced
  search space.
\newblock In {\em CVPR workshops}, pages 702--703, 2020.

\bibitem{PaCo}
Jiequan Cui, Zhisheng Zhong, Shu Liu, Bei Yu, and Jiaya Jia.
\newblock Parametric contrastive learning.
\newblock In {\em ICCV}, pages 715--724, 2021.

\bibitem{CB}
Yin Cui, Menglin Jia, Tsung-Yi Lin, Yang Song, and Serge Belongie.
\newblock Class-balanced loss based on effective number of samples.
\newblock In {\em CVPR}, pages 9268--9277, 2019.

\bibitem{BERT}
Jacob Devlin, Ming{-}Wei Chang, Kenton Lee, and Kristina Toutanova.
\newblock {BERT:} pre-training of deep bidirectional transformers for language
  understanding.
\newblock In {\em NAACL}, pages 4171--4186. Association for Computational
  Linguistics, 2019.

\bibitem{ViT}
Alexey Dosovitskiy, Lucas Beyer, Alexander Kolesnikov, Dirk Weissenborn,
  Xiaohua Zhai, Thomas Unterthiner, Mostafa Dehghani, Matthias Minderer, Georg
  Heigold, Sylvain Gelly, Jakob Uszkoreit, and Neil Houlsby.
\newblock An image is worth 16x16 words: Transformers for image recognition at
  scale.
\newblock In {\em ICLR}, 2021.

\bibitem{MASL}
Christoph Feichtenhofer, Haoqi Fan, Yanghao Li, and Kaiming He.
\newblock Masked autoencoders as spatiotemporal learners.
\newblock {\em arXiv preprint arXiv:2205.09113}, 2022.

\bibitem{ECE}
Chuan Guo, Geoff Pleiss, Yu Sun, and Kilian~Q Weinberger.
\newblock On calibration of modern neural networks.
\newblock In {\em ICML}, pages 1321--1330. PMLR, 2017.

\bibitem{MAE}
Kaiming He, Xinlei Chen, Saining Xie, Yanghao Li, Piotr Doll{\'{a}}r, and
  Ross~B. Girshick.
\newblock Masked autoencoders are scalable vision learners.
\newblock In {\em CVPR}, pages 15979--15988. {IEEE}, 2022.

\bibitem{MoCo}
Kaiming He, Haoqi Fan, Yuxin Wu, Saining Xie, and Ross Girshick.
\newblock Momentum contrast for unsupervised visual representation learning.
\newblock In {\em CVPR}, pages 9729--9738, 2020.

\bibitem{ResNet}
Kaiming He, Xiangyu Zhang, Shaoqing Ren, and Jian Sun.
\newblock Deep residual learning for image recognition.
\newblock In {\em CVPR}, pages 770--778, 2016.

\bibitem{DiVE}
Yin-Yin He, Jianxin Wu, Xiu-Shen Wei, et~al.
\newblock Distilling virtual examples for long-tailed recognition.
\newblock In {\em ICCV}, pages 235--244, 2021.

\bibitem{LADE}
Youngkyu Hong, Seungju Han, Kwanghee Choi, Seokjun Seo, Beomsu Kim, and Buru
  Chang.
\newblock Disentangling label distribution for long-tailed visual recognition.
\newblock In {\em CVPR}, pages 6626--6636, 2021.

\bibitem{SAFA}
Yan Hong, Jianfu Zhang, Zhongyi Sun, and Ke Yan.
\newblock Safa: Sample-adaptive feature augmentation for long-tailed image
  classification.
\newblock In {\em ECCV}, 2022.

\bibitem{BatchFormer}
Zhi Hou, Baosheng Yu, Dacheng Tao, et~al.
\newblock Batchformer: Learning to explore sample relationships for robust
  representation learning.
\newblock In {\em CVPR}, 2022.

\bibitem{CBD}
Ahmet Iscen, Andre Araujo, Boqing Gong, and Cordelia Schmid.
\newblock Class-balanced distillation for long-tailed visual recognition.
\newblock In {\em BMVC}, page 165. {BMVA} Press, 2021.

\bibitem{DAP}
Muhammad~Abdullah Jamal, Matthew Brown, Ming-Hsuan Yang, Liqiang Wang, and
  Boqing Gong.
\newblock Rethinking class-balanced methods for long-tailed visual recognition
  from a domain adaptation perspective.
\newblock In {\em CVPR}, pages 7610--7619, 2020.

\bibitem{KCL}
Bingyi Kang, Yu Li, Sa Xie, Zehuan Yuan, and Jiashi Feng.
\newblock Exploring balanced feature spaces for representation learning.
\newblock In {\em ICLR}, 2020.

\bibitem{EBFS}
Bingyi Kang, Yu Li, Sa Xie, Zehuan Yuan, and Jiashi Feng.
\newblock Exploring balanced feature spaces for representation learning.
\newblock In {\em ICLR}, 2021.

\bibitem{NCM}
Bingyi Kang, Saining Xie, Marcus Rohrbach, Zhicheng Yan, Albert Gordo, Jiashi
  Feng, and Yannis Kalantidis.
\newblock Decoupling representation and classifier for long-tailed recognition.
\newblock In {\em ICLR}, 2020.

\bibitem{SCL}
Prannay Khosla, Piotr Teterwak, Chen Wang, Aaron Sarna, Yonglong Tian, Phillip
  Isola, Aaron Maschinot, Ce Liu, and Dilip Krishnan.
\newblock Supervised contrastive learning.
\newblock {\em NeurIPS}, 33:18661--18673, 2020.

\bibitem{M2m}
Jaehyung Kim, Jongheon Jeong, Jinwoo Shin, et~al.
\newblock M2m: Imbalanced classification via major-to-minor translation.
\newblock In {\em CVPR}, pages 13896--13905, 2020.

\bibitem{Cifar}
A. Krizhevsky and G. Hinton.
\newblock Learning multiple layers of features from tiny images.
\newblock {\em Master's thesis, Department of Computer Science, University of
  Toronto}, 2009.

\bibitem{TLC}
Bolian Li, Zongbo Han, Haining Li, Huazhu Fu, and Changqing Zhang.
\newblock Trustworthy long-tailed classification.
\newblock In {\em CVPR}, pages 6970--6979, 2022.

\bibitem{NCL}
Jun Li, Zichang Tan, Jun Wan, Zhen Lei, and Guodong Guo.
\newblock Nested collaborative learning for long-tailed visual recognition.
\newblock In {\em CVPR}, pages 6949--6958, 2022.

\bibitem{GCL}
Mengke Li, Yiu-ming Cheung, Yang Lu, et~al.
\newblock Long-tailed visual recognition via gaussian clouded logit adjustment.
\newblock In {\em CVPR}, pages 6929--6938, 2022.

\bibitem{TSC}
Tianhong Li, Peng Cao, Yuan Yuan, Lijie Fan, Yuzhe Yang, Rogerio~S Feris, Piotr
  Indyk, and Dina Katabi.
\newblock Targeted supervised contrastive learning for long-tailed recognition.
\newblock In {\em CVPR}, pages 6918--6928, 2022.

\bibitem{SSD}
Tianhao Li, Limin Wang, and Gangshan Wu.
\newblock Self supervision to distillation for long-tailed visual recognition.
\newblock In {\em ICCV}, pages 630--639, 2021.

\bibitem{Focal}
Tsung-Yi Lin, Priya Goyal, Ross Girshick, Kaiming He, and Piotr Doll{\'a}r.
\newblock Focal loss for dense object detection.
\newblock In {\em ICCV}, pages 2980--2988, 2017.

\bibitem{COCO}
Tsung-Yi Lin, Michael Maire, Serge Belongie, James Hays, Pietro Perona, Deva
  Ramanan, Piotr Doll{\'a}r, and C~Lawrence Zitnick.
\newblock Microsoft coco: Common objects in context.
\newblock In {\em ECCV}, pages 740--755. Springer, 2014.

\bibitem{Bread}
Bo Liu, Haoxiang Li, Hao Kang, Gang Hua, and Nuno Vasconcelos.
\newblock Breadcrumbs: Adversarial class-balanced sampling for long-tailed
  recognition.
\newblock In {\em ECCV}, 2022.

\bibitem{rwSAM}
Hong Liu, Jeff~Z. HaoChen, Adrien Gaidon, and Tengyu Ma.
\newblock Self-supervised learning is more robust to dataset imbalance.
\newblock In {\em ICLR}, 2022.

\bibitem{TokenMix}
Jihao Liu, Boxiao Liu, Hang Zhou, Hongsheng Li, and Yu Liu.
\newblock Tokenmix: Rethinking image mixing for data augmentation in vision
  transformers.
\newblock In {\em ECCV}, 2022.

\bibitem{Swin}
Ze Liu, Yutong Lin, Yue Cao, Han Hu, Yixuan Wei, Zheng Zhang, Stephen Lin, and
  Baining Guo.
\newblock Swin transformer: Hierarchical vision transformer using shifted
  windows.
\newblock In {\em ICCV}, 2021.

\bibitem{OLTR}
Ziwei Liu, Zhongqi Miao, Xiaohang Zhan, Jiayun Wang, Boqing Gong, and Stella~X.
  Yu.
\newblock Large-scale long-tailed recognition in an open world.
\newblock In {\em CVPR}, 2019.

\bibitem{RAC}
Alexander Long, Wei Yin, Thalaiyasingam Ajanthan, Vu Nguyen, Pulak Purkait,
  Ravi Garg, Alan Blair, Chunhua Shen, and Anton van~den Hengel.
\newblock Retrieval augmented classification for long-tail visual recognition.
\newblock In {\em CVPR}, pages 6959--6969, 2022.

\bibitem{AdamW}
Ilya Loshchilov and Frank Hutter.
\newblock Decoupled weight decay regularization.
\newblock In {\em ICLR}, 2019.

\bibitem{LA}
Aditya~Krishna Menon, Sadeep Jayasumana, Ankit~Singh Rawat, Himanshu Jain,
  Andreas Veit, and Sanjiv Kumar.
\newblock Long-tail learning via logit adjustment.
\newblock In {\em ICLR}, 2021.

\bibitem{CKT}
Sarah Parisot, Pedro~M Esperan{\c{c}}a, Steven McDonagh, Tamas~J Madarasz,
  Yongxin Yang, and Zhenguo Li.
\newblock Long-tail recognition via compositional knowledge transfer.
\newblock In {\em CVPR}, pages 6939--6948, 2022.

\bibitem{CMO}
Seulki Park, Youngkyu Hong, Byeongho Heo, Sangdoo Yun, and Jin~Young Choi.
\newblock The majority can help the minority: Context-rich minority
  oversampling for long-tailed classification.
\newblock In {\em CVPR}, pages 6887--6896, 2022.

\bibitem{IB}
Seulki Park, Jongin Lim, Younghan Jeon, and Jin~Young Choi.
\newblock Influence-balanced loss for imbalanced visual classification.
\newblock In {\em ICCV}, pages 735--744, 2021.

\bibitem{BS}
Jiawei Ren, Cunjun Yu, Xiao Ma, Haiyu Zhao, Shuai Yi, et~al.
\newblock Balanced meta-softmax for long-tailed visual recognition.
\newblock {\em NeurIPS}, 33:4175--4186, 2020.

\bibitem{Imagenet}
Olga Russakovsky, Jia Deng, Hao Su, Jonathan Krause, Sanjeev Satheesh, Sean Ma,
  Zhiheng Huang, Andrej Karpathy, Aditya Khosla, Michael Bernstein,
  Alexander~C. Berg, and Li Fei-Fei.
\newblock {ImageNet Large Scale Visual Recognition Challenge}.
\newblock {\em IJCV}, 115(3):211--252, 2015.

\bibitem{EQL}
Jingru Tan, Changbao Wang, Buyu Li, Quanquan Li, Wanli Ouyang, Changqing Yin,
  and Junjie Yan.
\newblock Equalization loss for long-tailed object recognition.
\newblock In {\em CVPR}, pages 11662--11671, 2020.

\bibitem{CausalNorm}
Kaihua Tang, Jianqiang Huang, and Hanwang Zhang.
\newblock Long-tailed classification by keeping the good and removing the bad
  momentum causal effect.
\newblock {\em NeurIPS}, 33:1513--1524, 2020.

\bibitem{deit3}
Hugo Touvron, Matthieu Cord, and Herv{\'e} J{\'e}gou.
\newblock Deit iii: Revenge of the vit.
\newblock In {\em ECCV}, 2022.

\bibitem{FixRes}
Hugo Touvron, Andrea Vedaldi, Matthijs Douze, and Herv{\'e} J{\'e}gou.
\newblock Fixing the train-test resolution discrepancy.
\newblock In {\em NeurIPS}, 2019.

\bibitem{iNat}
Grant Van~Horn, Oisin Mac~Aodha, Yang Song, Yin Cui, Chen Sun, Alex Shepard,
  Hartwig Adam, Pietro Perona, and Serge Belongie.
\newblock The inaturalist species classification and detection dataset.
\newblock In {\em CVPR}, pages 8769--8778, 2018.

\bibitem{DOC}
Hualiang Wang, Siming Fu, Xiaoxuan He, Hangxiang Fang, Zuozhu Liu, and Haoji
  Hu.
\newblock Towards calibrated hyper-sphere representation via distribution
  overlap coefficient for long-tailed learning.
\newblock In {\em ECCV}, 2022.

\bibitem{RSG}
Jianfeng Wang, Thomas Lukasiewicz, Xiaolin Hu, Jianfei Cai, and Zhenghua Xu.
\newblock {RSG:} {A} simple but effective module for learning imbalanced
  datasets.
\newblock In {\em CVPR}, pages 3784--3793. Computer Vision Foundation / {IEEE},
  2021.

\bibitem{HybirdSC}
Peng Wang, Kai Han, Xiu-Shen Wei, Lei Zhang, and Lei Wang.
\newblock Contrastive learning based hybrid networks for long-tailed image
  classification.
\newblock In {\em CVPR}, pages 943--952, 2021.

\bibitem{RIDE}
Xudong Wang, Long Lian, Zhongqi Miao, Ziwei Liu, and Stella~X. Yu.
\newblock Long-tailed recognition by routing diverse distribution-aware
  experts.
\newblock In {\em ICLR}. OpenReview.net, 2021.

\bibitem{CreST}
Chen Wei, Kihyuk Sohn, Clayton Mellina, Alan Yuille, and Fan Yang.
\newblock Crest: A class-rebalancing self-training framework for imbalanced
  semi-supervised learning.
\newblock In {\em CVPR}, pages 10857--10866, 2021.

\bibitem{iNatFinegrained}
Xiu{-}Shen Wei, Peng Wang, Lingqiao Liu, Chunhua Shen, and Jianxin Wu.
\newblock Piecewise classifier mappings: Learning fine-grained learners for
  novel categories with few examples.
\newblock {\em {IEEE} Trans. Image Process.}, 28(12):6116--6125, 2019.

\bibitem{iNatFG}
Xiu-Shen Wei, Peng Wang, Lingqiao Liu, Chunhua Shen, and Jianxin Wu.
\newblock Piecewise classifier mappings: Learning fine-grained learners for
  novel categories with few examples.
\newblock {\em IEEE Transactions on Image Processing}, 28(12):6116--6125, 2019.

\bibitem{timm}
Ross Wightman, Hugo Touvron, and Herv{\'e} J{\'e}gou.
\newblock Resnet strikes back: An improved training procedure in timm.
\newblock {\em arXiv preprint arXiv:2110.00476}, 2021.

\bibitem{DBL}
Tong Wu, Qingqiu Huang, Ziwei Liu, Yu Wang, and Dahua Lin.
\newblock Distribution-balanced loss for multi-label classification in
  long-tailed datasets.
\newblock In {\em ECCV}, pages 162--178. Springer, 2020.

\bibitem{LFME}
Liuyu Xiang, Guiguang Ding, Jungong Han, et~al.
\newblock Learning from multiple experts: Self-paced knowledge distillation for
  long-tailed classification.
\newblock In {\em ECCV}, pages 247--263. Springer, 2020.

\bibitem{ResNeXt}
Saining Xie, Ross Girshick, Piotr Doll{\'a}r, Zhuowen Tu, and Kaiming He.
\newblock Aggregated residual transformations for deep neural networks.
\newblock In {\em CVPR}, pages 1492--1500, 2017.

\bibitem{DLSA}
Yue Xu, Yong-Lu Li, Jiefeng Li, and Cewu Lu.
\newblock Constructing balance from imbalance for long-tailed image
  recognition.
\newblock In {\em ECCV}, pages 38--56. Springer, 2022.

\bibitem{PriorLT}
Zhengzhuo Xu, Zenghao Chai, Chun Yuan, et~al.
\newblock Towards calibrated model for long-tailed visual recognition from
  prior perspective.
\newblock {\em NeurIPS}, 34:7139--7152, 2021.

\bibitem{SSP}
Yuzhe Yang, Zhi Xu, et~al.
\newblock Rethinking the value of labels for improving class-imbalanced
  learning.
\newblock {\em NeurIPS}, 33:19290--19301, 2020.

\bibitem{ID-samlping}
Sihao Yu, Jiafeng Guo, Ruqing Zhang, Yixing Fan, Zizhen Wang, and Xueqi Cheng.
\newblock A re-balancing strategy for class-imbalanced classification based on
  instance difficulty.
\newblock In {\em CVPR}, pages 70--79, 2022.

\bibitem{Cutmix}
Sangdoo Yun, Dongyoon Han, Seong~Joon Oh, Sanghyuk Chun, Junsuk Choe, and
  Youngjoon Yoo.
\newblock Cutmix: Regularization strategy to train strong classifiers with
  localizable features.
\newblock In {\em ICCV}, pages 6023--6032, 2019.

\bibitem{mixup}
Hongyi Zhang, Moustapha Cisse, Yann~N. Dauphin, and David Lopez-Paz.
\newblock mixup: Beyond empirical risk minimization.
\newblock In {\em ICLR}, 2018.

\bibitem{DisAlign}
Songyang Zhang, Zeming Li, Shipeng Yan, Xuming He, and Jian Sun.
\newblock Distribution alignment: A unified framework for long-tail visual
  recognition.
\newblock In {\em CVPR}, pages 2361--2370, 2021.

\bibitem{Range}
Xiao Zhang, Zhiyuan Fang, Yandong Wen, Zhifeng Li, and Yu Qiao.
\newblock Range loss for deep face recognition with long-tailed training data.
\newblock In {\em ICCV}, pages 5409--5418, 2017.

\bibitem{TADE}
Yifan Zhang, Bryan Hooi, Lanqing Hong, and Jiashi Feng.
\newblock Test-agnostic long-tailed recognition by test-time aggregating
  diverse experts with self-supervision.
\newblock {\em arXiv preprint arXiv:2107.09249}, 2021.

\bibitem{Bagoftricks}
Yongshun Zhang, Xiu-Shen Wei, Boyan Zhou, and Jianxin Wu.
\newblock Bag of tricks for long-tailed visual recognition with deep
  convolutional neural networks.
\newblock In {\em AAAI}, pages 3447--3455, 2021.

\bibitem{ALA}
Yan Zhao, Weicong Chen, Xu Tan, Kai Huang, and Jihong Zhu.
\newblock Adaptive logit adjustment loss for long-tailed visual recognition.
\newblock In {\em AAAI}, volume~36, pages 3472--3480, 2022.

\bibitem{MiSLAS}
Zhisheng Zhong, Jiequan Cui, Shu Liu, and Jiaya Jia.
\newblock Improving calibration for long-tailed recognition.
\newblock In {\em CVPR}, pages 16489--16498. Computer Vision Foundation /
  {IEEE}, 2021.

\bibitem{BBN}
Boyan Zhou, Quan Cui, Xiu-Shen Wei, and Zhao-Min Chen.
\newblock Bbn: Bilateral-branch network with cumulative learning for
  long-tailed visual recognition.
\newblock In {\em CVPR}, pages 9719--9728, 2020.

\bibitem{PlaceLT}
Bolei Zhou, Agata Lapedriza, Aditya Khosla, Aude Oliva, and Antonio Torralba.
\newblock Places: A 10 million image database for scene recognition.
\newblock {\em IEEE TPAMI}, 2017.

\bibitem{BCL}
Jianggang Zhu, Zheng Wang, Jingjing Chen, Yi-Ping~Phoebe Chen, and Yu-Gang
  Jiang.
\newblock Balanced contrastive learning for long-tailed visual recognition.
\newblock In {\em CVPR}, pages 6908--6917, 2022.

\end{thebibliography}
    }
}

\newpage
\onecolumn
\appendix

\begin{center}
    \Large \textbf{Learning Imbalanced Data with Vision Transformers}
    \Large \\ \textbf{Supplementary Material}
\end{center}
\vspace{20pt}

\section{Missing Proofs and Derivations}

\subsection{Proof to Theorem 1}
\noindent\textbf{Theorem 1.} \hl{\normalfont{{Logit Bias of Balanced CE.}}} Let $\pi_{\mathbf{y}_i}=n_{\mathbf{y}_i} / N$ be the training label $\mathbf{y}_i$ distribution. If we implement the balanced cross-entropy loss via logit adjustment, the bias item of logit 
$\mathbf{z}_{\mathbf{y}_i}$ will be $\mathcal{B}^{\text{ce}}_{\mathbf{y}_i}=\log \pi_{\mathbf{y}_i}$, \textit{i.e.},
\begin{equation*}
  \begin{aligned}
        \mathcal{L}_{\text{Bal-CE}} = \log[1+\sum_{\mathbf{y}_j\neq \mathbf{y}_i} e^{(\mathbf{z}_{\mathbf{y}_j} + \log \pi_{\mathbf{y}_j}) - (\mathbf{z}_{\mathbf{y}_i} + \log \pi_{\mathbf{y}_i})}].
\end{aligned}  
\end{equation*}

\noindent\textbf{\textit{Proof.}}
\\

Following the notions in Section Preliminaries, we simplify a model $\mathcal{M}_{\theta}$ with parameters $\theta$, which attempts to learn the joint probability distribution of images and labels $\mathcal{P}(\mathcal{X}, \mathcal{Y})$. Due to its agnostic, one may try to get the maximum posterior $\mathcal{P}(\mathcal{Y}|\mathcal{X})$ as an approximation solution from the Bayesian estimation view. To this end, if we Maximize A Posterior (MAP) to optimize $\theta$, we have:
\begin{equation*}
  \begin{aligned}
  \hat{\theta} = \arg \max_{\theta} \mathcal{P}(\mathcal{Y}|\mathcal{X}) = \arg \max_{\theta} \frac{\mathcal{P}(\mathcal{X}|\mathcal{Y}) \cdot \mathcal{P}(\mathcal{Y})}{\mathcal{P}(\mathcal{X})} = \arg \max_{\theta} \mathcal{P}(\mathcal{X}|\mathcal{Y}) \cdot \mathcal{P}(\mathcal{Y}),
  \end{aligned}  
\end{equation*}
where $\mathcal{P}(\mathcal{X}|\mathcal{Y})$ is the likelihood function, $\mathcal{P}(\mathcal{Y})$ is the prior distribution of $\mathcal{Y}$, and $\mathcal{P}(\mathcal{X})$ is the evidence factor, which is $\theta$ irrelevant. Then, if we reasonably view $\mathcal{P}(\mathcal{Y})$ as the class distribution (typically class label frequency $\pi_{\mathbf{y}_i}$ as approximations), the MAP is equivalent to maximizing the likelihood function $\mathcal{P}(\mathcal{X}|\mathcal{Y}; \theta)$. Considering both training $\mathcal{P}^{s}(\mathcal{X},\mathcal{Y})$ and test datasets $\mathcal{P}^{t}(\mathcal{X},\mathcal{Y})$, the MAP shall hold on to both of them, \textit{i.e.},
\begin{equation*}
\left\{\begin{aligned}
  & \hat{\theta} = \arg \max_{\theta} \mathcal{P}^{s}(\mathcal{Y}|\mathcal{X}) =  \arg \max_{\theta} \mathcal{P}^{s}(\mathcal{X}|\mathcal{Y};\theta) \cdot \mathcal{P}^{s}(\mathcal{Y}) \\
  & \hat{\theta} = \arg \max_{\theta} \mathcal{P}^{t}(\mathcal{Y}|\mathcal{X}) =  \arg \max_{\theta} \mathcal{P}^{t}(\mathcal{X}|\mathcal{Y};\theta) \cdot \mathcal{P}^{t}(\mathcal{Y})
  \end{aligned}\right. 
\end{equation*}

With model parameters $\theta$ learned on the training set $\mathcal{P}^{s}(\mathcal{X},\mathcal{Y})$, the likelihood function will be consistent. To obtain the maximization posterior on the test dataset (the best accuracy performance), we can derive that:
\begin{equation*}
\begin{aligned}
  \mathcal{P}^{t}(\mathcal{Y}|\mathcal{X}; \theta) \quad \propto \quad \mathcal{P}^{t}(\mathcal{X}|\mathcal{Y};\theta) \mathcal{P}^{t}(\mathcal{Y}) \quad \propto \quad \frac{\mathcal{P}^{s}(\mathcal{Y}|\mathcal{X};\theta)}{\mathcal{P}^{s}(\mathcal{Y})} \cdot \mathcal{P}^{t}(\mathcal{Y})
\end{aligned}
\end{equation*}

Since MAP is equivalent to maximizing the likelihood function $\mathcal{P}(\mathcal{X}|\mathcal{Y}; \theta)$, we further decouple the test MAP as regulation terms to achieve the Structural Risk Minimization:
\begin{equation*}
\begin{aligned}
  \arg \max_{\theta} \mathcal{P}^{t}(\mathcal{Y}|\mathcal{X}; \theta)\  = \  \arg \max_{\theta} \log \mathcal{P}^{t}(\mathcal{Y}|\mathcal{X}; \theta) \  = \  \arg \max_{\theta} \log \mathcal{P}^{s}(\mathcal{X}|\mathcal{Y};\theta) - \log \mathcal{P}^{s}(\mathcal{Y}) + \log \mathcal{P}^{t}(\mathcal{Y})
\end{aligned}
\end{equation*}

Notice that $\mathcal{P}^{s}(\mathcal{Y})$ and $\mathcal{P}^{t}(\mathcal{Y})$ are both $\theta$ irrelevant according to our previous hypothesis. Hence, we can compensate the regulation terms $- \log \mathcal{P}^{s}(\mathcal{Y}) + \log \mathcal{P}^{t}(\mathcal{Y})$ during the training procession as $+ \log \pi^{s}(\mathbf{y}) - \log \pi^{t}(\mathbf{y})$. In addition, if we adopt the \textit{Softmax} for probability normalization, we will have:
\begin{equation*}
\begin{aligned}
    \mathcal{P}^{s}(\mathbf{x}_i|\mathbf{y}_i;\theta) = \frac{e^{\mathbf{z}_{\mathbf{y}_i}}}{\sum_{\mathbf{y}_j \in \mathcal{C}} e^{\mathbf{z}_{\mathbf{y}_j}}} \quad \Longrightarrow \quad \log \mathcal{P}^{s}(\mathbf{x}_i|\mathbf{y}_i;\theta) = \log \frac{e^{\mathbf{z}_{\mathbf{y}_i}}}{\sum_{\mathbf{y}_j \in \mathcal{C}} e^{\mathbf{z}_{\mathbf{y}_j}}} \quad \propto \quad \log e^{\mathbf{z}_{\mathbf{y}_i}} = \mathbf{z}_{\mathbf{y}_i}
\end{aligned}
\end{equation*}

Thus, the $\log \mathcal{P}^{s}(\mathcal{X}|\mathcal{Y};\theta)$ is equivalent to the output logits $\mathbf{z} := \mathcal{M}(\mathbf{x}|\theta)$ and we immediately deduce that the training regulation shall be $\mathbf{z}_{\mathbf{y}} + \log \pi^{s}_\mathbf{y} - \log \pi^{t}_\mathbf{y}$. For the balanced test datasets, $- \log \pi^{t}_\mathbf{y}=\log C$ and can be ignored for all classes. Hence, we derive the final bias as:
\begin{equation*}
\begin{aligned}
    \mathcal{B}^{\text{ce}}_{\mathbf{y}_i}=\log \pi^{s}_{\mathbf{y}_i}
\end{aligned}
\end{equation*}

\qed
\clearpage

\subsection{Proof to Theorem 2\&3}
\label{sec:proof_thm2}
\noindent\textbf{Theorem 2\&3.} \normalfont{\hl{Logit Bias of Balanced BCE with Test Prior.}} Let $\pi^s_{\mathbf{y}_i}$ and $\pi^t_{\mathbf{y}_i}$ be the label $\mathbf{y}_i$ training and test distribution. If we implement the balanced cross-entropy loss via logit adjustment, the bias item of logit $\mathbf{z}_{\mathbf{y}_i}$ will be: 
\begin{equation*}
\begin{aligned}
\mathcal{B}^{\text{bce}}_{\mathbf{y}_i}=(\log \pi^s_{\mathbf{y}_i} - \log \pi^t_{\mathbf{y}_i}) - (\log (1-\pi^s_{\mathbf{y}_i}) - \log (1-\pi^t_{\mathbf{y}_i}))
\end{aligned}
\end{equation*}
\noindent\textbf{\textit{Proof.}}
\\

In this paper, we propose the balanced binary cross entropy loss in Thm.~\ref{thm:02_balance_sigmoid} and further extend it with the test prior (test label distribution) in Thm.~\ref{thm:03_balance_sigmoid_test}. As we discussed, the bias in Thm.~\ref{thm:02_balance_sigmoid} is derived from  re-balancing with training instance numbers like \cite{BS} do. Here, we give another proof from the Bayesian estimation view like Thm.~\ref{thm:01_balance_softmax}. We mainly give the proof to the Thm.~\ref{thm:03_balance_sigmoid_test} and derive the Thm.~\ref{thm:02_balance_sigmoid} as a special case of Thm.~\ref{thm:03_balance_sigmoid_test}. Following the notions in the proof to Thm.~\ref{thm:01_balance_softmax}, BCE loss treats the long-tailed recognition task as $C$ independent binary classification problems. For every single problem, the derivation in Thm.~\ref{thm:01_balance_softmax} still holds if $\mathcal{Y}:=\{0,1\}$:
\begin{equation*}
\begin{aligned}
  \arg \max_{\theta} \mathcal{P}^{t}(\mathcal{Y}|\mathcal{X}; \theta) \ = \  \arg \max_{\theta} \log \mathcal{P}^{t}(\mathcal{Y}|\mathcal{X}; \theta) \ = \  \arg \max_{\theta} \log \mathcal{P}^{s}(\mathcal{X}|\mathcal{Y};\theta) - \log \mathcal{P}^{s}(\mathcal{Y}) + \log \mathcal{P}^{t}(\mathcal{Y})
\end{aligned}
\end{equation*}

If we adopt the \textit{Sigmoid} for probability normalization, we will have:
\begin{equation*}
\begin{aligned}
    \mathcal{P}^{s}(\mathbf{x}_i|\mathbf{y}_i;\theta) = \frac{1}{1+e^{{-\mathbf{z}_{\mathbf{y}_i}}}} \quad \Longrightarrow \quad \frac{e^{{\mathbf{z}_{\mathbf{y}_i}}}}{e^{{\mathbf{z}_{\mathbf{y}_i}}} + e^{0}}
\end{aligned}
\end{equation*}

Similar to the \textit{Softmax}, for the binary classification, we consider the ${e^{{\mathbf{z}_{\mathbf{y}_i}}}} / {(e^{{\mathbf{z}_{\mathbf{y}_i}}} + e^{0})}$ as the likelihood for $\mathcal{Y}=1$ and ${e^{0}} / {(e^{{\mathbf{z}_{\mathbf{y}_i}}} + e^{0})}$ for $\mathcal{Y}=0$. Then, we can derive that:

\begin{equation*}
\begin{aligned}
    \log \mathcal{P}^{s}(\mathbf{x}_i|\mathbf{y}_i;\theta) = \log \frac{e^{{\mathbf{z}_{\mathbf{y}_i}}}}{e^{{\mathbf{z}_{\mathbf{y}_i}}} + e^{0}} \quad \propto \quad \log e^{\mathbf{z}_{\mathbf{y}_i}} = \mathbf{z}_{\mathbf{y}_i}
\end{aligned}
\end{equation*}

Different from CE, which just punishes the positive term, BCE shall take the negative terms into consideration as well. If we take the statistical label frequency $\pi^{s}_{\mathbf{y}}$ and $\pi^{t}_{\mathbf{y}}$ as the prior, we can deduce that the bias should be:
\begin{equation*}
\left\{\begin{aligned}
    & \log \pi^{s}_\mathbf{y} - \log \pi^{t}_\mathbf{y} \quad \quad \quad \quad \quad \quad \quad \text{\textit{for positive item} $\mathbf{z}_{\mathbf{y}_i}$} \\
    & \log (1-\pi^{s}_\mathbf{y}) - \log (1- \pi^{t}_\mathbf{y}) \quad \quad \ \text{\textit{for negative item} $0$}
\end{aligned}\right.
\end{equation*}

Hence, for a single binary classification, the unbiased \textit{Sigmoid} operation is required to compensate for each term:
\begin{equation*}
\begin{aligned}
    \sigma(\mathbf{z}_{\mathbf{y}_i}) = \frac{e^{{\mathbf{z}_{\mathbf{y}_i}}}}{e^{{\mathbf{z}_{\mathbf{y}_i}}} + e^{0}} \Longrightarrow \frac{e^{{\mathbf{z}_{\mathbf{y}_i}}+\log \pi^{s}_{\mathbf{y}_i} - \log \pi^{t}_{\mathbf{y}_i}}}{e^{{\mathbf{z}_{\mathbf{y}_i}}+\log \pi^{s}_{\mathbf{y}_i} - \log \pi^{t}_{\mathbf{y}_i}} + e^{0+\log (1-\pi^{s}_{\mathbf{y}_i}) - \log (1- \pi^{t}_{\mathbf{y}_i})}}
\end{aligned}
\end{equation*}

To match the Logit Adjustment requirement \cite{LA}, we convert all bias to the logit $\mathbf{z}_{\mathbf{y}_i}$:
\begin{equation*}
\begin{aligned}
    \sigma(\mathbf{z}_{\mathbf{y}_i}) = \frac{e^{{\mathbf{z}_{\mathbf{y}_i}}}}{e^{{\mathbf{z}_{\mathbf{y}_i}}} + e^{0}} &\Longrightarrow \frac{e^{{\mathbf{z}_{\mathbf{y}_i}}+\log \pi^{s}_{\mathbf{y}_i} - \log \pi^{t}_{\mathbf{y}_i}-(\log (1-\pi^{s}_{\mathbf{y}_i}) - \log (1- \pi^{t}_{\mathbf{y}_i}))}}{e^{{\mathbf{z}_{\mathbf{y}_i}}+\log \pi^{s}_{\mathbf{y}_i} - \log \pi^{t}_{\mathbf{y}_i}-(\log (1-\pi^{s}_{\mathbf{y}_i}) - \log (1- \pi^{t}_{\mathbf{y}_i}))} + e^{0}} \\
    & = \frac{1}{1 + e^{-[{\mathbf{z}_{\mathbf{y}_i}}+ (\log \pi^{s}_{\mathbf{y}_i} - \log \pi^{t}_{\mathbf{y}_i})-(\log (1-\pi^{s}_{\mathbf{y}_i}) - \log (1- \pi^{t}_{\mathbf{y}_i}))]}} 
\end{aligned}
\end{equation*}

Hence, we get the final bias with train and test label prior knowledge:
\begin{equation*}
\begin{aligned}
\mathcal{B}^{\text{bce}}_{\mathbf{y}_i}=(\log \pi^s_{\mathbf{y}_i} - \log \pi^t_{\mathbf{y}_i}) - (\log (1-\pi^s_{\mathbf{y}_i}) - \log (1-\pi^t_{\mathbf{y}_i}))
\end{aligned}
\end{equation*}

For the balanced test dataset, $\pi^t_{\mathbf{y}_i}=1/C$ and the $\mathcal{B}^{\text{bce}}_{\mathbf{y}_i}$ will be the form in Thm.~\ref{thm:02_balance_sigmoid} if we ignore constant terms.
\begin{equation*}
\begin{aligned}
\mathcal{B}^{\text{bce}}_{\mathbf{y}_i}=(\log \pi^s_{\mathbf{y}_i} - \log \frac{1}{C}) - (\log (1-\pi^s_{\mathbf{y}_i}) - \log (1-\frac{1}{C})) = \log \pi^s_{\mathbf{y}_i} - \log (1-\pi^s_{\mathbf{y}_i}) + \log (C-1)
\end{aligned}
\end{equation*}

\qed
\clearpage

\subsection{Fisher Consistency with Test Prior}
Menon \textit{et al.} show how to verify whether a pair-wise loss ensures Fisher consistency for the balanced error (see the Theorem 1 in \cite{LA}). Here, we extend it to test the prior available situations. \\
\begin{equation*}
\begin{aligned}
\mathcal{L}(\mathbf{y}_i, \mathcal{M}(\mathbf{x}))=\alpha_{\mathbf{y}_i} \cdot \log [1+\sum_{\mathbf{y}_j \neq \mathbf{y}_i} Exp(\Delta_{\mathbf{y}_i \mathbf{y}_j}) \cdot Exp(\mathcal{M}_{\mathbf{y}_j}(\mathbf{x})-\mathcal{M}_{\mathbf{y}_i}(\mathbf{x}))]
\end{aligned}
\end{equation*}

\noindent\textbf{Theorem 4.} For any $\delta^{{s}}, \delta^{{t}} \in \mathbb{R}_{+}^{{C}}$, the pairwise loss is Fisher consistent with weights and margins:
$$
\alpha_{\mathbf{y}_{i}}=\frac{\delta_{\mathbf{y}_{i}}^{s} \cdot \pi_{\mathbf{y}_{i}}^{t}}{\delta_{\mathbf{y}_{i}}^{t} \cdot \pi_{\mathbf{y}_{i}}^{s}} \quad \Delta_{\mathbf{y}_{i} \mathbf{y}_{j}}=\log \left(\frac{\delta_{\mathbf{y}_{j}}^{s} \cdot \delta_{\mathbf{y}_{i}}^{t}}{\delta_{\mathbf{y}_{j}}^{t} \cdot \delta_{\mathbf{y}_{i}}^{s}}\right)
$$

With $\delta_{\mathbf{y}_{i}}^{s}=\pi_{\mathbf{y}_{i}}^{s}$ and $\delta_{\mathbf{y}_{i}}^{t}=\pi_{\mathbf{y}_{i}}^{t}$, we deduce that Bal-BCE is Fisher consistent between train (s) and test (t) set.
\\

\noindent\textbf{\textit{Proof.}}

Let $\Delta_{{\mathbf{y}_i}{\mathbf{y}_j}}=\log \left(\frac{\delta_{\mathbf{y}_j}^{s} \cdot \delta_{\mathbf{y}_i}^{t}}{\delta_{\mathbf{y}_j}^{t} \cdot \delta_{\mathbf{y}_i}^{s}}\right)$ and $\alpha_{\mathbf{y}_i}=1$, we have:
$$
\mathcal{L}(\mathbf{y}_i, \mathcal{M}(\mathbf{x}))=-\log \frac{{e}^{\mathbf{z}_{\mathbf{y}_{i}}+\log \delta_{\mathbf{y}_{i}}^{s}-\log \delta_{\mathbf{y}_{i}}^{t}}}{\sum_{\mathbf{y}_{j} \in \mathcal{Y}} e^{\mathbf{z}_{\mathbf{y}_j} + \log \delta_{\mathbf{y}_{j}}^{s}-\log \delta_{\mathbf{y}_{j}}^{t}}}
$$

If $\eta_{\mathbf{y}_{i}}(\mathbf{x})$ represents the posterior possibility $\mathcal{P}^{s}(\mathbf{y}_{i}|\mathbf{x})$, the Bayes-optimal score will satisfy:
$$
\mathbf{z}^*_{\mathbf{y}_{i}} + \log \delta_{\mathbf{y}_{i}}^{s} - \log \delta_{\mathbf{y}_{i}}^{t} = \log \eta_{\mathbf{y}_{i}}(\mathbf{x}) \quad \Longrightarrow \quad \mathbf{z}^*_{\mathbf{y}_{i}} = \log \left(\frac{\eta_{\mathbf{y}_{i}}(\mathbf{x})}{\delta_{\mathbf{y}_{i}}^{s}} \cdot \delta_{\mathbf{y}_{i}}^{t} \right)
$$

Now consider adding weights $\alpha_{\mathbf{y}_{i}}$ to the loss term, the corresponding risk shall be:
$$
\mathbb{E}_{\mathbf{x}, \mathbf{y}}\left[\mathcal{L}_{\alpha_{\mathbf{y}_{i}}}\right] \quad = \quad \sum_{\mathbf{y}_{i} \in \mathcal{Y}} \pi_{\mathbf{y}_{i}}^{s} \cdot \mathbb{E}_{\mathbf{x} \mid \mathbf{y}=\mathbf{y}_{i}}\left[\mathcal{L}_{\alpha_{\mathbf{y}_{i}}}\right] \quad = \quad \sum_{\mathbf{y}_{i} \in \mathcal{Y}} \pi_{\mathbf{y}_{i}}^{s} \cdot \alpha_{\mathbf{y} i} \cdot \mathbb{E}_{\mathbf{x} \mid \mathbf{y}=\mathbf{y}_{i}}[\mathcal{L}] \quad \propto \quad \sum_{\mathbf{y}_{i} \in \mathcal{Y}} \bar{\pi}_{\mathbf{y}_{i}}^{s} \cdot \mathbb{E}_{\mathbf{x} \mid \mathbf{y}=\mathbf{y}_{i}}[\mathcal{L}]
$$

where $\bar{\pi}_{\mathbf{y}_{i}}^{s} \propto \pi_{\mathbf{y}_{i}}^{s} \cdot \alpha_{\mathbf{y}_{i}}$. Hence training with the weighted loss amounts to training with the original loss on the new label distribution $\bar{\pi}$. The posterior probability $\bar{\eta}_{\mathbf{y}_{i}}(\mathbf{x})$ on the altered label distribution is:
$$
\bar{\eta}_{\mathbf{y}_{i}}(\mathbf{x}) =\overline{\mathcal{P}}\left(\mathbf{y}_{i}|\mathbf{x}\right) \quad \propto \quad \mathcal{P}\left(\mathbf{x}|\mathbf{y}_{i}\right) \cdot \bar{\pi}_{\mathbf{y}_{i}}^{s} \quad \propto \quad \eta_{\mathbf{y}_{i}}(\mathbf{x}) \cdot \frac{\bar{\pi}_{\mathbf{y}_{i}}^{\mathrm{s}}}{\pi_{\mathbf{y}_{i}}^{s}} \quad \propto \quad \eta_{\mathbf{y}_{i}}(\mathbf{x}) \cdot \alpha_{\mathbf{y}_{i}}
$$

When we set $\alpha_{\mathbf{y}_{i}}=\frac{\delta_{\mathbf{y}_{i}}^{s} \cdot \pi_{\mathbf{y}_{i}}^{t}}{\delta_{\mathbf{y}_{i}}^{t} \cdot \pi_{\mathbf{y}_{i}}^{s}}$, the Bayes-optimal score will satisfy:
\begin{equation*}
\begin{aligned}
\arg \max_{\mathbf{y}_{i} \in \mathcal{Y}} \mathbf{z}_{\mathbf{y}_{i}}^{*} \quad &= \quad \arg \max_{\mathbf{y}_{i} \in \mathcal{Y}} \log \left(\frac{\bar{\eta}_{\mathbf{\mathbf{y}_{i}}}(\mathbf{x})}{\delta_{\mathbf{y}_{i}}^{s}} \cdot \delta_{\mathbf{y}_{i}}^{t}\right) \quad \\ &= \quad \arg \max_{\mathbf{y}_{i} \in \mathcal{Y}}  \log \left(\frac{\eta_{\mathbf{y}_{i}}(\mathbf{x}) \cdot \alpha_{\mathbf{y}_{i}}}{\delta_{\mathbf{y}_{\mathbf{i}}}^{s}} \cdot \delta_{\mathbf{y}_{i}}^{t}\right) \quad \\ &= \quad \arg \max_{\mathbf{y}_{i} \in \mathcal{Y}}  \log \left(\frac{\eta_{\mathbf{y}_{i}}(\mathbf{x})}{\pi_{\mathbf{y}_{i}}^{s}} \cdot \pi_{\mathbf{y}_{i}}^{t}\right) 
\end{aligned}
\end{equation*}

\qed
\clearpage

\section{Analysis to Proposed Bias}
\begin{wrapfigure}{r}{0.4\textwidth}
\vspace{-20pt}
\includegraphics[width=0.4\textwidth]{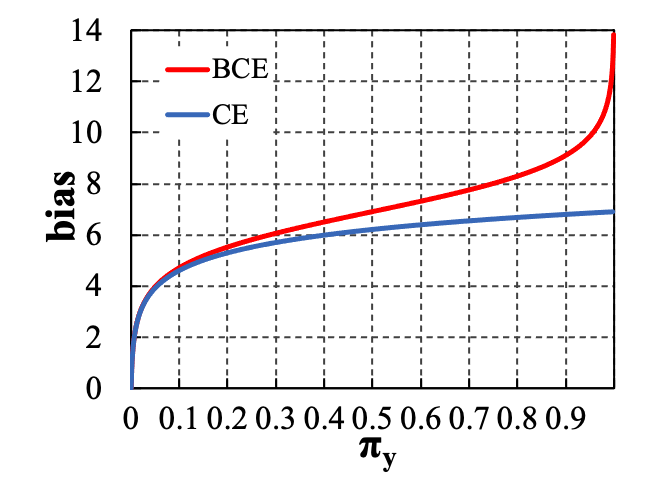}
\caption{$\mathcal{B}^{\text{ce}}_{\mathbf{y}_i}$ and $\mathcal{B}^{\text{bce}}_{\mathbf{y}_i}$ \textit{w.r.t.} $\pi_{\mathbf{y}_i}$ ($C$=1,000).}
\vspace{-10pt}
\label{fig:val-ce-bce}
\end{wrapfigure}
For Bal-CE, Ren \textit{et al.} \cite{BS} propose the balanced softmax as a strong baseline for long-tailed recognition while Menon \textit{et al} \cite{LA} deploy it by adding extra logit margins. The following works \cite{LADE, PriorLT} further extend it with test prior knowledge, which can be written as:
\begin{equation*}
    \mathcal{B}^{\text{ce}}_{\mathbf{y}_i} = \log \pi^s_{\mathbf{y}_i} + \log C
\end{equation*}

To improve the performance of balanced binary cross-entropy loss in long-tailed recognition, we propose an unbiased version of \textit{Sigmoid} to eliminate the inherent bias to the head class. Inspired by Logit Adjustment \cite{LA}, we implement it as a bias $\mathcal{B}^{\text{bce}}_{\mathbf{y}_i}$ to the model logits and extend to test prior as well, which can be written as:
\begin{equation*}
    \mathcal{B}^{\text{bce}}_{\mathbf{y}_i} = \log \pi^s_{\mathbf{y}_i} - \log (1-\pi^s_{\mathbf{y}_i}) + \log(C-1)
\end{equation*}

Fig.~\ref{fig:val-ce-bce} shows the difference between $\mathcal{B}^{\text{ce}}_{\mathbf{y}_i}$ and $\mathcal{B}^{\text{bce}}_{\mathbf{y}_i}$. Notice that $\mathcal{B}^{\text{bce}}_{\mathbf{y}_i}$ is closed to $\mathcal{B}^{\text{ce}}_{\mathbf{y}_i}$ when $\pi_{\mathbf{y}_i}$ is small, which indicates that both $\mathcal{B}^{\text{ce}}_{\mathbf{y}_i}$ and $\mathcal{B}^{\text{bce}}_{\mathbf{y}_i}$ help the models to pay more attention to learn the tail. However, $\mathcal{B}^{\text{bce}}_{\mathbf{y}_i}$ gives larger biases to the head and makes the inter-class distance of the head smaller. Such a modification allows Bal-BCE to show more tolerance to the head compared to Bal-CE. To be more specific, CE utilizes \textit{Softmax} to emphasize mutual exclusion, where large head bias will damage corresponding performance severely. In contrast, BCE calculates independent class-wise probability with \textit{Sigmoid} function, where the original task is considered as a series of binary classification tasks. Hence, the head bias will not influence the tail. In addition, larger biases will not hurt the head as CE does because it hedges the over-suppression for negative labels. CE can not benefit from it because of its mutual exclusion. 

\begin{equation*}
\left\{\begin{aligned}
& \frac{\partial \mathcal{L}_{\text{Bal-CE}}\left(\mathbf{z}_{\mathbf{y}_j}, \mathbbm{1}({\mathbf{y}_j})\right)}{\partial\left(\mathbf{z}_{\mathbf{y}_j}\right)} = \frac{e^{\mathbf{z}_{\mathbf{y}_j}+\mathcal{B}^{\text{ce}}_{\mathbf{y}_j}}}{\sum_{\mathbf{y}_i \in \mathcal{C}} e^{\mathbf{z}_{\mathbf{y}_i} + \mathcal{B}^{\text{ce}}_{\mathbf{y}_i}}}, \quad \frac{\partial \mathcal{L}_{\text{Bal-BCE}}\left(\mathbf{z}_{\mathbf{y}_j}, \mathbbm{1}({\mathbf{y}_j})\right)}{\partial\left(\mathbf{z}_{\mathbf{y}_j}\right)} = \frac{e^{\mathbf{z}_{\mathbf{y}_j} + \mathcal{B}^{\text{bce}}_{\mathbf{y}_j}}}{1+e^{\mathbf{z}_{\mathbf{y}_j} + \mathcal{B}^{\text{bce}}_{\mathbf{y}_j}}}, \quad \ \ \ \mathbbm{1}({\mathbf{y}_j})=0 \\
& \frac{\partial \mathcal{L}_{\text{Bal-CE}}\left(\mathbf{z}_{\mathbf{y}_j}, \mathbbm{1}({\mathbf{y}_j})\right)}{\partial\left(\mathbf{z}_{\mathbf{y}_j}\right)} = \frac{e^{\mathbf{z}_{\mathbf{y}_j}+\mathcal{B}^{\text{ce}}_{\mathbf{y}_j}}}{\sum_{\mathbf{y}_i \in \mathcal{C}} e^{\mathbf{z}_{\mathbf{y}_i} + \mathcal{B}^{\text{ce}}_{\mathbf{y}_i}}}, \quad \frac{\partial \mathcal{L}_{\text{Bal-BCE}}\left(\mathbf{z}_{\mathbf{y}_j}, \mathbbm{1}({\mathbf{y}_j})\right)}{\partial\left(\mathbf{z}_{\mathbf{y}_j}\right)} = -\frac{1}{1+e^{\mathbf{z}_{\mathbf{y}_j} + \mathcal{B}^{\text{bce}}_{\mathbf{y}_j}}}, \quad \mathbbm{1}({\mathbf{y}_j})=1
\end{aligned}\right.
\end{equation*}

From the optimization view, as the above equation shows, we can also observe that $\mathcal{B}^{\text{bce}}_{\mathbf{y}_i}$ will not affect class $\mathbf{y}_j$'s gradients. However, for Bal-CE, the optimization step would be rather small once the logit for the positive class is much higher than those of the negative ones. With the dominance of head labels, larger head biases will make the networks fall into even worse situations. In contrast, for the Bal-BCE, the above larger head biases will act as a regularization to overcome the over-suppression while avoiding damage to the head classes themselves.

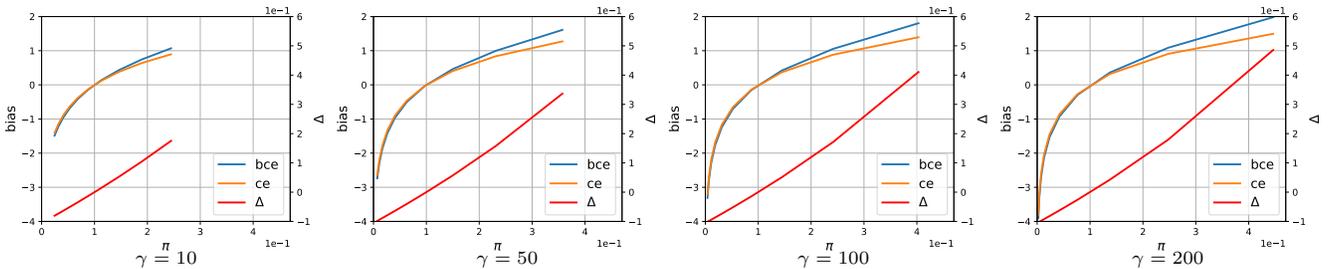
\begin{figure*}[!b]
\centering
\begin{overpic}[scale=0.35,grid=False,tics=5]{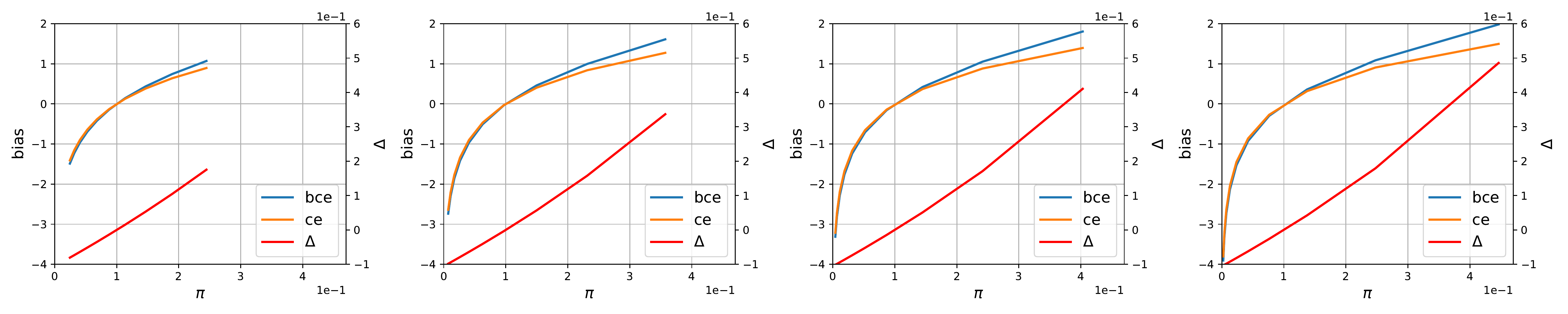}
\put(10.5,0){\scriptsize{$\gamma=10$}}
\put(36,0){\scriptsize{$\gamma=50$}}
\put(60,0){\scriptsize{$\gamma=100$}}
\put(85,0){\scriptsize{$\gamma=200$}}
\end{overpic}
\caption{Visualization of the bias in CIFAR10-LT dataset. A larger $\gamma$ indicates a severer imbalance situation. $\Delta$ is the difference between the two biases, which is shown in right \textit{y}-axis. With $\gamma$ increases, the $\Delta$ becomes more important to the final bias.}
\label{fig:vis-bias}
\end{figure*}

In addition, $\mathcal{B}^{\text{bce}}_{\mathbf{y}_i}$ will be more important when the datasets become more skewed. As Fig.~\ref{fig:vis-bias} shows, the difference will be larger when the imbalance factor $\gamma$ increases. It means the performance will get worse if we adopt $\mathcal{B}^{\text{ce}}_{\mathbf{y}_i}$ for BCE loss. Notice that the gap between $\mathcal{B}^{\text{ce}}_{\mathbf{y}_i}$ and $\mathcal{B}^{\text{bce}}_{\mathbf{y}_i}$ has consistent diminution when the class number $C$ is getting bigger. However, $\mathcal{B}^{\text{bce}}_{\mathbf{y}_i}$ still bring obvious performance gain in
this circumstance.

\clearpage

\section{Datasets}
We conduct experiments on CIFAR-LT \cite{Cifar}, ImageNet-LT \cite{Imagenet}, iNat18 \cite{iNat}, and Places-LT \cite{PlaceLT}. With different imbalanced factors $\gamma$, we build the long-tailed version of CIFAR by discarding training instances following the rule given in \cite{CB} and keeping the original validation set for all datasets. To investigate the MGP performance on LT data, we build a balanced ImageNet-1K subset called ImageNet-BAL. It contains the same training instance number as ImageNet-LT while keeping class labels balanced. Notice that both LT and BAL adopt the same validation set. We demonstrate MGP is robust enough for long-tailed data via quantitative and qualitative experiments on the BAL and LT. iNat18 is the largest benchmark of the long tail community. Our \name ameliorates vanilla ViTs most significantly because of the data scale and fine-grained problems. Places-LT is created from large-scale dataset Places \cite{PlaceLT} by \cite{OLTR}. The train set contains just 62K images with a high imbalance factor, which makes it challenging for data-hungry Transformers.

\begin{table}[ht]
\centering
\caption{Detailed information of datasets motioned in the main paper.}
\begin{tabular}{@{}c|cccc|c|c|c|c@{}}
\toprule
\multirow{3}{*}{Dataset} & \multicolumn{2}{c|}{CIFAR-10-LT}     & \multicolumn{2}{c|}{CIFAR-100-LT} & \multirow{3}{*}{ImageNet-LT} & \multirow{3}{*}{ImageNet-BAL} & \multirow{3}{*}{iNat18} & \multirow{3}{*}{PlaceLT} \\ \cmidrule(lr){2-5}
                         & \multicolumn{4}{c|}{Imbalance Factor ($\gamma$)}                                               &                              &                               &                         &                          \\ \cmidrule(lr){2-5}
                         & 100    & \multicolumn{1}{c|}{10}     & 100             & 10              &                              &                               &                         &                          \\ \midrule
Training Images          & 12,406 & \multicolumn{1}{c|}{20,431} & 10,847          & 19,573          & 115,846                      & 160,000                       & 437,513                 & 62,500                   \\
Classes Number           & 10     & \multicolumn{1}{c|}{10}     & 100             & 100             & 1,000                        & 1,000                         & 8,142                   & 365                      \\
Max Images               & 5,000  & \multicolumn{1}{c|}{5,000}  & 500             & 500             & 1,280                        & 160                           & 1,000                   & 4,980                    \\
Min Images               & 50     & \multicolumn{1}{c|}{500}    & 5               & 50              & 5                            & 160                           & 2                       & 5                        \\
Imbalance Factor                    & 100    & \multicolumn{1}{c|}{10}     & 100             & 10              & 256                          & 1                             & 500                     & 996                      \\ \bottomrule
\end{tabular}
\end{table}

\section{Implementation Details}
\subsection{Augmentations in Algorithm.1}  
In Alg.~\ref{alg:pipeline}, \name adopts different augmentations in two stages, \textit{i.e.}, $\mathcal{A}_{pt}$ \& $\mathcal{A}_{ft}$. The reason is from our observations that the strong data augmentations in MGP will not contribute to higher performance while bringing extra calculation burden. Some augmentations like Color Jitter may lead to wired reconstruction results \textit{w.r.t.} the augmented images. For the BFT stage, we adopt more general data augmentations for stable training procession. The AutoAug improves performance on ImageNet-LT/BAL remarkably and slightly in iNat18 / Places-LT, which is consistent with the observation in \cite{PaCo}. Mixup and Cutmix make the training more smooth, and RandomErease regulates the model with better performance.

\begin{table}[ht]
\centering
\caption{The detailed augmentations adopted in Alg.~\ref{alg:pipeline}.}
\begin{tabular}{@{}c|cc@{}}
\toprule
Augmentation         & Masked Generative Pretraining ($\mathcal{A}_{pt}$) & Balanced Fine Tuning ($\mathcal{A}_{ft}$) \\ \midrule
RandomResizedCrop    & \checkmark         & \checkmark           \\
RandomHorizontalFlip & \checkmark         & \checkmark           \\
AutoAug              & $\times$           & (9,0.5)              \\
Mixup                & $\times$           & 0.8                  \\
Cutmix               & $\times$           & 1.0                  \\
RandomErease         & $\times$           & 0.25                 \\
Normalize            & \checkmark         & \checkmark           \\ \bottomrule
\end{tabular}
\label{tab:augmentations}
\end{table}

\subsection{Configure Settings for Table 1}
In Tab.~\ref{tab:pretrain-recips}, we implement different ViT training recipes on long-tailed and balanced ImageNet-1K subsets. Specially, we reproduce vanilla ViTs according to Tab.11 in \cite{MAE}, DeiT III according to Tab.1 in \cite{deit3}, and MAE according to Tab.9 in \cite{MAE}. All recipes train ViTs with more epochs (800) compared to ResNets (typically 90 or 180). However, the performance is far from catching up with ResNet baselines and severely deteriorate when it becomes imbalanced because the dataset is relatively small for data-hungry ViTs compared to ImageNet-1K or ImageNet22K and the long-tailed labels bias the ViTs heavily.

\subsection{Configure Settings for the Main Comparisons}
We conduct experiments on ImageNet-LT, iNat18, and Places-LT. For fair comparisons, we train all models from scratch following previous LTR work. To balance the performance and computation complexity trade-off, we adopt a small image size for the large-scale dataset and adopt 800 epochs for MGP. Thanks to the masked tokens, MGP trains ViTs much faster than vanilla ViT and DeiT. We transfer the hyper-parameters of ImageNet-LT to other benchmarks and just finetune the $\tau$ of Bal-BCE loss slightly. Notice that Places-LT is a small dataset and we just finetune 30 epochs to avoid over-fitting.

\begin{table}[h]
\centering
\caption{The \name configurations on three main benchmarks. We mainly transfer the hyper-parameters of ImageNet-LT to other benchmarks without wide changes. Tuning hyper-parameters will bring further improvement.}
\setlength{\tabcolsep}{7mm}
\begin{tabular}{@{}cccc@{}}
\toprule
\multicolumn{1}{c|}{Configuration}            & ImageNet-LT      & iNaturalist 2018          & Places-LT            \\ \hline
\multicolumn{4}{c}{\cellcolor{lightgrey}Masked Generative Pretraining.} \\ \hline
\multicolumn{1}{c|}{Epoch}                & 800              & 800              & 800              \\
\multicolumn{1}{c|}{Warmup Epoch}         & 40               & 40               & 40               \\
\multicolumn{1}{c|}{Effective Batch Size} & 4096             & 4096             & 4096             \\
\multicolumn{1}{c|}{Optimizer}            & AdamW(0.9,0.95)  & AdamW(0.9,0.95)  & AdamW(0.9,0.95)  \\
\multicolumn{1}{c|}{Learning Rate}        & 1.5e-4         & 1.5e-4         & 1.5e-4         \\
\multicolumn{1}{c|}{LR schedule}          & cosine(min=0)    & cosine(min=0)    & cosine(min=0)    \\
\multicolumn{1}{c|}{Weight Decay}         & 5e-2         & 5e-2         & 5e-2         \\
\multicolumn{1}{c|}{Mask Ratio}           & 0.75             & 0.75             & 0.75             \\
\multicolumn{1}{c|}{Input Size}           & 224              & 128              & 224              \\ \hline
\multicolumn{4}{c}{\cellcolor{lightgrey}Balanced Fine Tuning.} \\ \hline
\multicolumn{1}{c|}{Epoch}                & 100              & 100              & 30               \\
\multicolumn{1}{c|}{Warmup Epoch}         & 10               & 10               & 5                \\
\multicolumn{1}{c|}{Effective Batch Size} & 1024             & 1024             & 1024             \\
\multicolumn{1}{c|}{Optimizer}            & AdamW(0.9,0.99)  & AdamW(0.9,0.99)  & AdamW(0.9,0.99)  \\
\multicolumn{1}{c|}{Learning Rate}        & 1e-3         & 1e-3         & 1e-3         \\
\multicolumn{1}{c|}{LR schedule}          & cosine(min=1e-6) & cosine(min=1e-6) & cosine(min=1e-6) \\
\multicolumn{1}{c|}{Weight Decay}         & 5e-2         & 5e-2         & 5e-2         \\
\multicolumn{1}{c|}{Layer Decay}          & 0.75             & 0.75             & 0.75             \\
\multicolumn{1}{c|}{Input Size}           & 224              & 224              & 224              \\
\multicolumn{1}{c|}{Drop Path}            & 0.1              & 0.2              & 0.1              \\
\multicolumn{1}{c|}{$\tau$ of Bal-BCE}       & 1                & 1                & 1.05             \\ \bottomrule
\end{tabular}
\label{tab:exp_settings}
\end{table}

\section{Additional Experiments}

\subsection{DeiT with Bal-BCE}
\label{sec:deit_balbce}
In the DeiT III \cite{deit3}, Touvron \textit{et al.} propose to train ViTs with binary cross entropy loss. With our proposed bias $\mathcal{B}^{\text{bce}}$, we can further boost its recipe when collaborating with long-tailed distributed data. As Tab.~\ref{tab:deit-bal} shows, Bal-BCE rebalances the performance of ViT-Small over three groups and improves the overall accuracy significantly. It is worth noticing that the few-shot gets ameliorated remarkably, while the many-shot is sacrificed to some extent. Compared to the results in Tab.\ref{tab:bias-ablation}, we get a meticulous observation that Bal-BCE improves all groups' performance when adopting MGP as the pretrain manner, and even the many-shot (head) classes get compelling growth, especially on the small models. The aforementioned phenomenon may indicate that the MGP learns more generalized and unbiased features compared to supervised manners, which helps $\mathcal{B}^{\text{bce}}$ to calibrate more misclassification cases instead of the over-confident but right cases.

\begin{table}[b]
\centering
\caption{Ablation study of proposed bias on DeiT III. Experiments are conducted with ViT-Small on ImageNet-LT for 400 epochs.}
\setlength{\tabcolsep}{5mm}
\begin{tabular}{@{}c|cc|cc|cc|cc@{}}
\toprule
Loss Type & Many & $\Delta$    & Med. & $\Delta$     & Few  & $\Delta$      & Acc  & $\Delta$     \\ \midrule
BCE w/o $\mathcal{B}^{\text{bce}}$ & 64.2 & -    & 32.2 & -    & 9.0  & -     & 41.4 & -    \\
BCE w/ $\mathcal{B}^{\text{bce}}$ & 60.3 & \bad{-4.0} & 40.8 & \good{+8.7} & 23.8 & \good{+14.7} & 46.0 & \good{+4.6} \\ \bottomrule
\end{tabular}
\label{tab:deit-bal}
\end{table}

\subsection{Performance with higher resolution}
With the FixRes effect \cite{FixRes}, LiVT can reach further performance gains with minor computational overhead, which only increases the resolution in the 2nd stage with a few epochs. As a comparison, ResNet-based methods require extra effort to modify the network with heavy computational overhead. Hence, we only provide LiVT$^*$ in Tab.~{\ref{tab:performance-imagenet-lt}-\ref{tab:performance-place}}. Note that LiVT with 224 resolution already achieves SOTA performance (except tiny Places-LT). We additionally show ViT-based methods with 384 resolution in Tab.~\ref{tab:resolution}. ViT-based methods typically show lower performance than ResNet-based ones due to ViTs' data hungry in the tiny dataset (Tab.~\ref{tab:performance-place}). Noteworthy, our Bal-BCE loss remarkably improves performance (Acc \textbf{+10.5}\% \& Few \textbf{+18.8}\% compared to MAE). While tuning hyper-parameters (e.g., $\tau$ in Alg.~\ref{alg:pipeline} or parameters in Tab.~\ref{tab:augmentations}\&\ref{tab:exp_settings}) can further boost the performance (Fig.~\ref{fig:hyp-tau}), we keep consistent settings with Tab.~\ref{tab:performance-imagenet-lt}\&\ref{tab:performance-iNat} to report the LiVT performance in Tab.~\ref{tab:performance-place}.

\begin{table}[h]
\centering
\setlength\tabcolsep{10pt}
\setlength{\tabcolsep}{5mm}
\caption{Top-1 Accuracy of ViT-B-16 pretrained on iNat18 dataset. We fine-tune models for 100 epochs with 384 resolution.}
\begin{tabular}{l|llll}
\toprule[1pt]
Resolution & ViT        & DeiT        & MAE         & LiVT     \\ \midrule[1pt]
224 $\times$ 224        & 54.6       & 61.0        & 69.4        & 76.1     \\
384 $\times$ 384      & 56.3 \good{+1.7} & 63.7 \good{+2.7} & 72.9 \good{+3.5} & 81.0 \good{+4.9} \\ \bottomrule[1pt]
\end{tabular}
\vspace{-5pt}
\label{tab:resolution}
\end{table}

\subsection{Negative-Tolerant Regularization}
Recently, there are some other works to improve the performance of BCE loss. For instance, Wu \textit{et al.} \cite{DBL} propose to leave more Negative Tolerant Regularization (NTR) in the BCE loss. In long-tailed recognition, the tail class samples are usually learned as negative pairs resulting from the head class dominance. Here, for clear and concise expression, we call the logit $z_{\mathbf{y}_i}$ positive logit and $z_{\mathbf{y}_j}, (j\neq i)$ negative logits for the label $\mathbf{y}_i$. For \textit{Softmax} operation, the gradient of the negative logits will be relatively small due to its mutual exclusion when the positive logit is large. However, \textit{Sigmoid} acts differently from \textit{Softmax}. The \textit{Sigmoid} always maintains relatively large gradients for negative logits despite the positive logit value. This property of BCE leads to the output tail class logits being smaller, which incurs that the model only overfits a few tail-positive samples in the training set.

To overcome this problem, Wu \textit{et al.} propose the NT-BCE loss to alleviate the dominance of negative labels. With a hyper-parameter $\lambda$ to control the strength of negative tolerance regularization, the NT-BCE can be written as:

\begin{equation*}
\begin{aligned}
    \mathcal{L}_{\text{NT-BCE}} = -\sum_{\mathbf{y}_i \in \mathcal{C}} &[\mathbbm{1}(\mathbf{y}_i) \cdot \log \frac{1}{1+e^{-\mathbf{z}_{\mathbf{y}_i}}} + \frac{1}{\lambda}(1-\mathbbm{1}(\mathbf{y}_i)) \cdot \log (1-\frac{1}{1+e^{-\lambda\mathbf{z}_{\mathbf{y}_i}}})]
\end{aligned}
\end{equation*}

To collaborate with it, we add our proposed bias $\mathcal{B}^{\text{bce}}_{\mathbf{y}_i} = \log \pi_{\mathbf{y}_i} - \log (1-\pi_{\mathbf{y}_i}))$ to the above loss and derive that:
\begin{equation*}
\begin{aligned}
    \mathcal{L}^*_{\text{NT-BCE}} = -\sum_{\mathbf{y}_i \in \mathcal{C}} &[\mathbbm{1}(\mathbf{y}_i) \cdot \log \frac{1}{1+e^{-(\mathbf{z}_{\mathbf{y}_i} + \mathcal{B}^{\text{bce}}_{\mathbf{y}_i})}} + \frac{1}{\lambda}(1-\mathbbm{1}(\mathbf{y}_i)) \cdot \log (1-\frac{1}{1+e^{-\lambda (\mathbf{z}_{\mathbf{y}_i} + \mathcal{B}^{\text{bce}}_{\mathbf{y}_i})}})]
\end{aligned}
\end{equation*}

\begin{figure*}[t]
\flushleft
\subfloat[(a) LT-10 with $\mathcal{L}_{\text{NT-BCE}}$]{
        \includegraphics[width=0.24\linewidth]{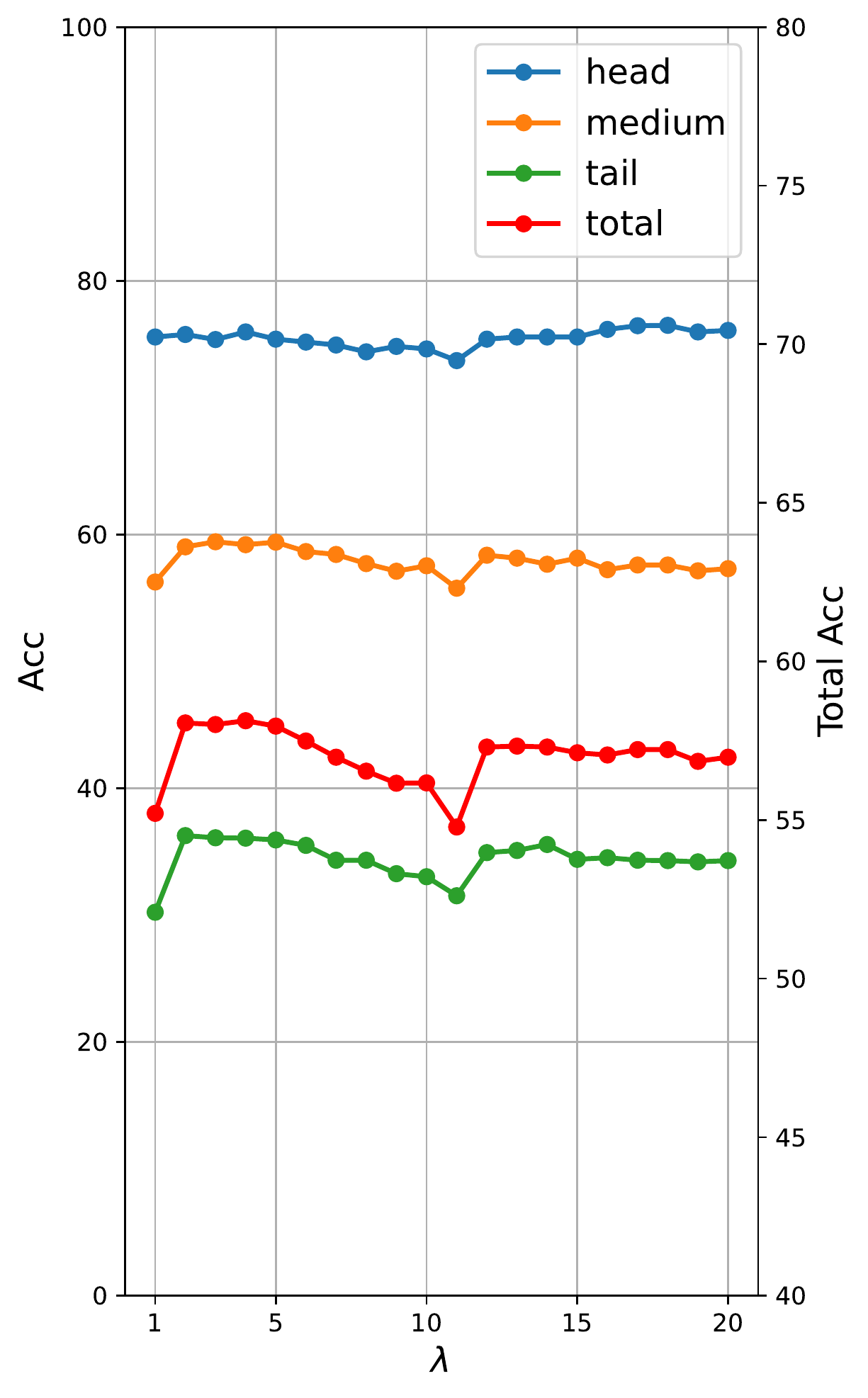}
    }
\subfloat[(b) LT-100 with $\mathcal{L}_{\text{NT-BCE}}$]{
        \includegraphics[width=0.24\linewidth]{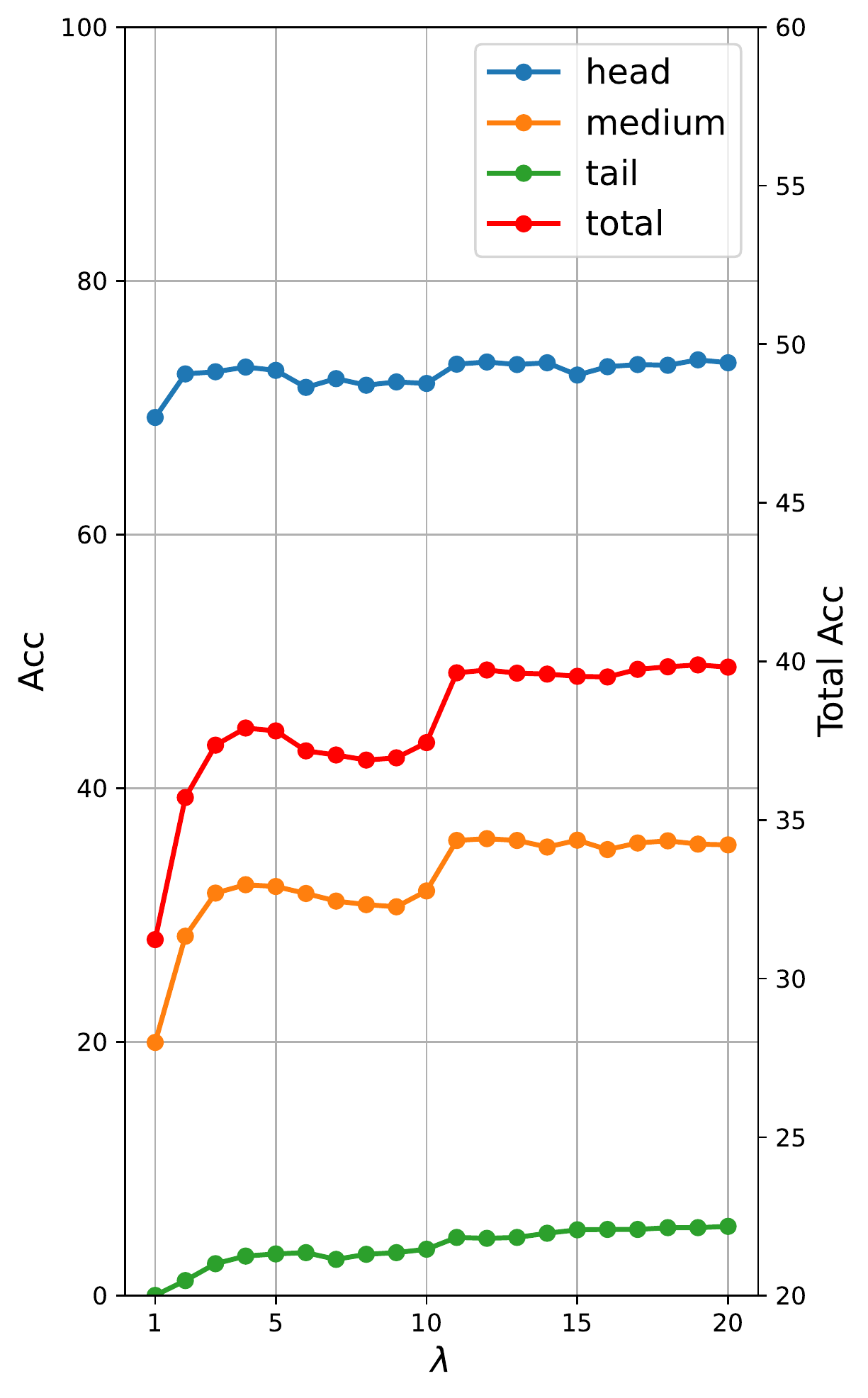}
    }
\subfloat[(c) LT-10 with $\mathcal{L}^*_{\text{NT-BCE}}$]{
        \includegraphics[width=0.24\linewidth]{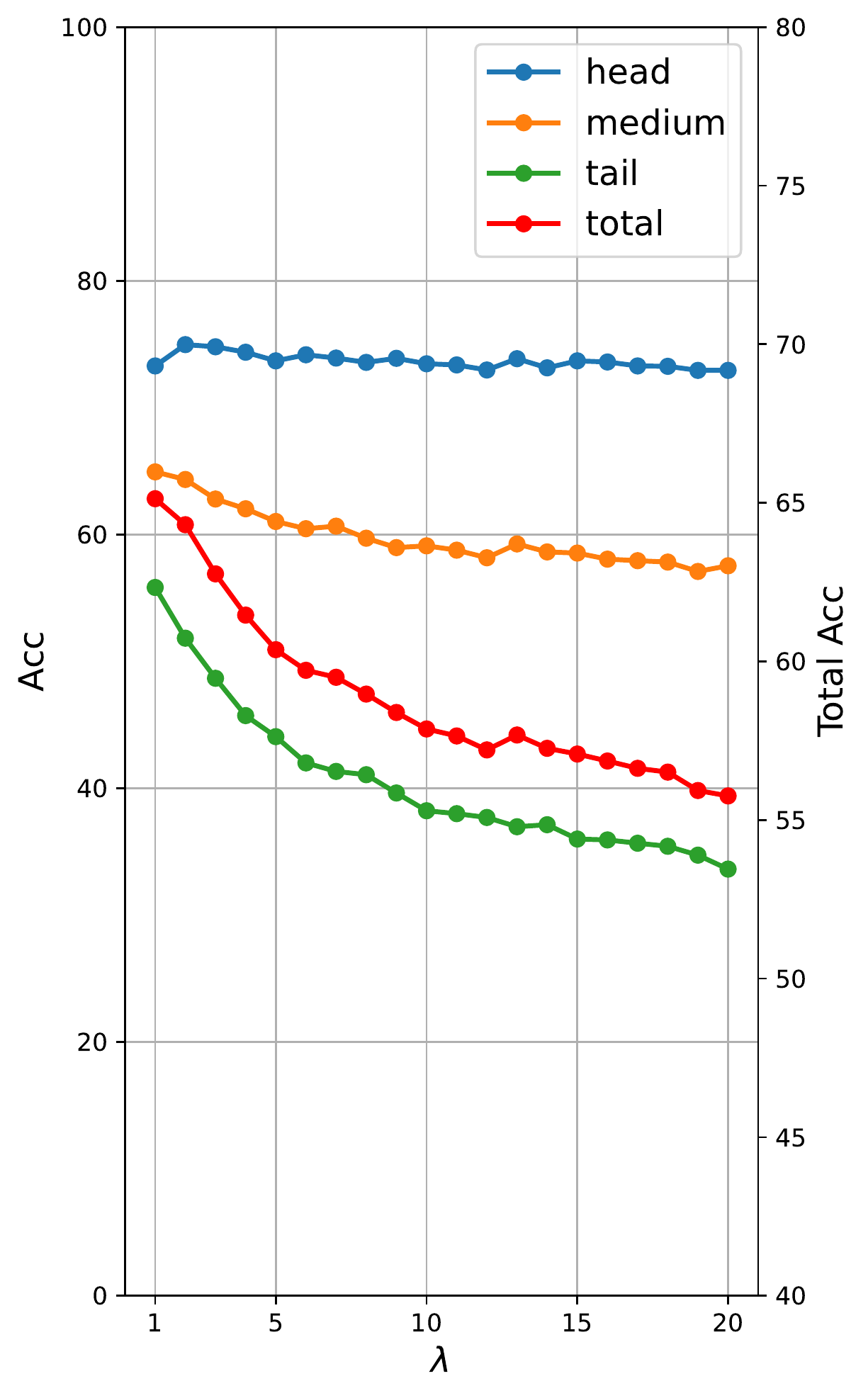}
    }
\subfloat[(d) LT-100 with $\mathcal{L}^*_{\text{NT-BCE}}$]{
        \includegraphics[width=0.24\linewidth]{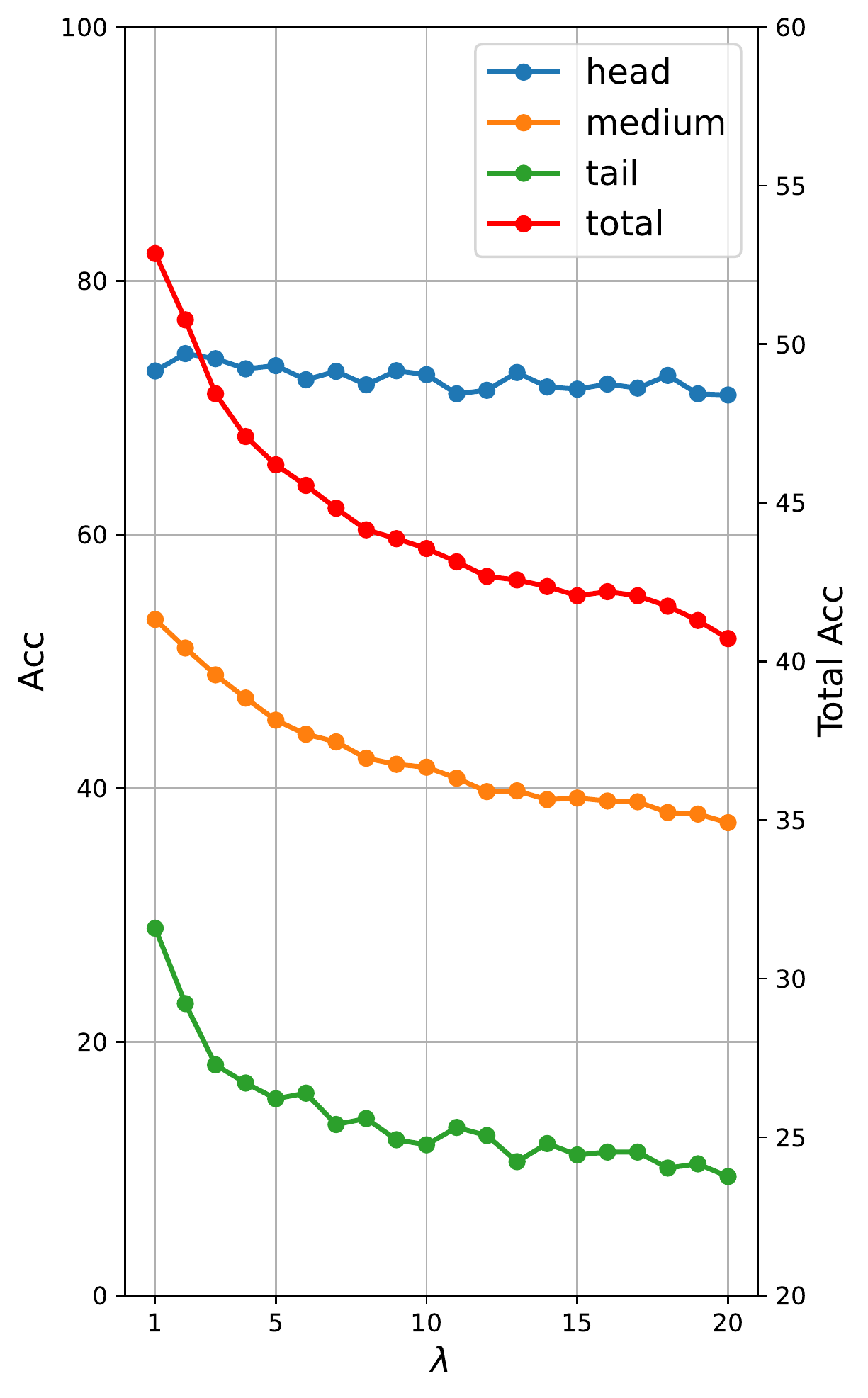}
    }
	\caption{Performance of BCE loss with NTR and $\mathcal{B}^{\text{bce}}$ on CIFAR100-LT. The total accuracy (red) is shown in right \textit{y}-axis for better visualizations. (a)(b) NTR boosts vanilla BCE loss by benefiting medium and tail classes. (c)(d) NTR fails to collaborate with our bias.}
\label{fig:ntr}
\end{figure*}

For more in-depth observations, we train ViT-B on CIFAT-100-LT with both $\mathcal{L}_{\text{NT-BCE}}$ and $\mathcal{L}^*_{\text{NT-BCE}}$ and show the experiment results in Fig.~\ref{fig:ntr}. The NTR ameliorates the vanilla BCE loss with large $\lambda$ by benefiting medium and tail classes. However, the performance of $\mathcal{L}_{\text{NT-BCE}}$ is hard to catch up with $\mathcal{L}^*_{\text{NT-BCE}}$. What's worse, the NTR consistently deteriorates the performance of $\mathcal{L}^*_{\text{NT-BCE}}$ when $\lambda$ gets larger. The best is achieved at $\lambda=1$, which indicates that NTR can not work well with our bias.

To explain it, we revisit the purpose of NTR, which aims to reduce the gradient of tail negative logits. While optimizing the tail class as negative logits, if the logit is small, the corresponding gradient will also be small to keep the logit from over-minimization. However, it is contradictory to our proposed bias. Typically, the margin-based loss makes the network pay attention to certain categories by increasing the corresponding difficulty with larger margins. As the margins for all classes, our bias $\mathcal{B}^{\text{bce}}$ makes the tail (head) class harder (easier) to learn, where the initial head logits are larger than tail ones, as shown in Fig.\ref{fig:vis-bias}. With NTR, tail classes will converge more slowly because larger $\lambda$ tends to slow down the optimization of tail logits, which finally results in unsatisfying tail performance. Although We \textit{et al.}, add a similar bias in \cite{DBL}, they ignore its effect because of the little difference between the training and test label distribution of their datasets. More explorations are still required to make NTR and $\mathcal{B}^{\text{bce}}$ complement each other in long-tailed recognition.

\captionsetup[subfloat]{labelsep=none,format=plain,labelformat=empty}

\section{More Discussions}

\subsection{About the two-stage pipeline.}
Although one stage is a promising direction, the two-stage frameworks (e.g., c-RT~\cite{NCM}, MiSLAS~\cite{MiSLAS}, and our LiVT) typically achieve much better performance. For ViTs in LTR tasks, the difficulty is to learn the inductive bias and label statistical bias simultaneously. We manage the challenge by decoupling the two biases and learning the inductive bias in the MGP stage and the statistical bias in the BFT stage separately.

\subsection{Test prior to model performance.}
The Bal-CE implementation in previous work~\cite{BS} contains the test prior (i.e., $\pi_i^t=1/C$) by default. With balanced test data, it is equal to eliminate the test prior bias item $-\log \pi_i^t=\log C$ with \textit{Softmax} operation (Eq.~\ref{eq:bal-ce}). However, for Bal-BCE, the test prior term cannot be reduced in \textit{Sigmoid} operation. As we discussed in Thm.~\ref{thm:03_balance_sigmoid_test}, although ignoring this term does not influence the optimization direction, it will reduce the loss value, especially when $C$ is large (c.f. derivation in Supp.~\ref{sec:proof_thm2}). Therefore, the test prior is essential to ensure stability during training in Bal-BCE (e.g., models trained in the iNat18 dataset cannot converge without this item).

\subsection{About the baseline performance.}
\begin{figure}[b!]
    \centering
    \includegraphics[width=0.8\linewidth]{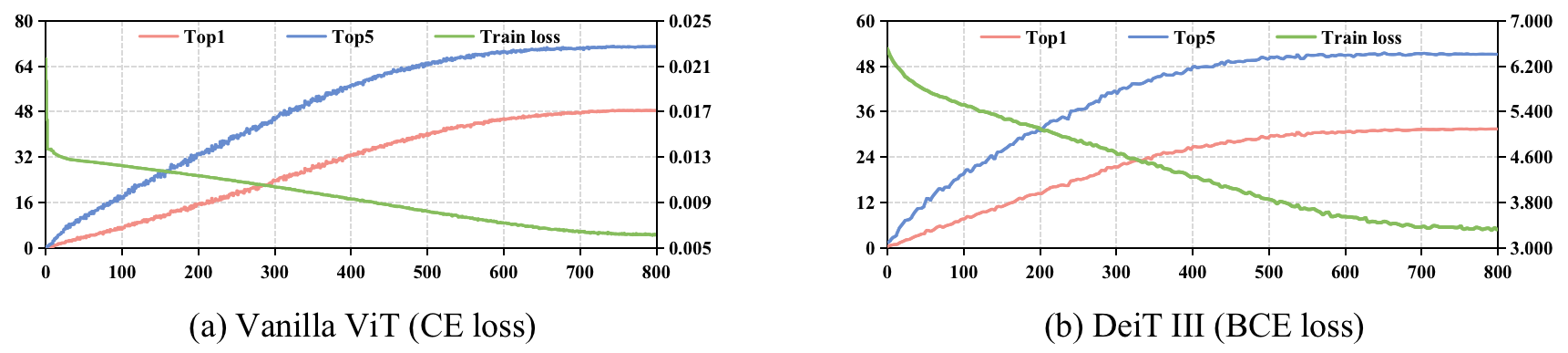}
    \caption{The training log of ViT-B on ImageNet-LT. Left y-axis: validation accuracy. Right y-axis: training loss value. The validation accuracy is consistent with training loss, which means that overfitting does not occur during the training process.}
\label{fig:baseline_overfit}
\end{figure}

One may consider overfitting as a possible reason for the poor performance of the ViT baseline. Hence, we visualize the training log in Fig.~\ref{fig:baseline_overfit}. Either in the tiny Places-LT or the large-scale iNat18 (similar scale to ImageNet-1K), ViTs exhibit biased performance (Tab.~{\ref{tab:performance-imagenet-lt}-\ref{tab:performance-place}}). The unsatisfactory performance of ViT-based baselines (direct supervision) mainly accounts for the long-tailed problems rather than the overfitting issues (Tab.~\ref{tab:pretrain-recips}). Even under the same setting with these baselines, Bal-BCE improves MAE (Tab.~{\ref{tab:performance-imagenet-lt}-\ref{tab:performance-place}}) and DeiT (Supp.~\ref{sec:deit_balbce}) consistently in few and overall performance.

\clearpage

\section{Visualization of MGP Reconstruction}
\captionsetup[subfloat]{labelsep=none,format=plain,labelformat=empty}
\begin{figure*}[h]
	\flushleft
	\subfloat[]{
        \includegraphics[width=0.98\linewidth]{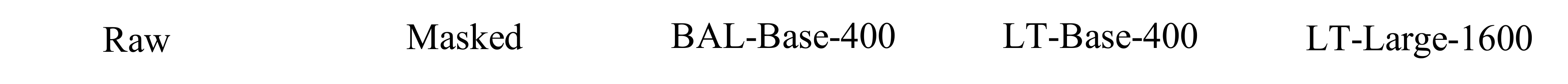}
    }
    \vspace{-15pt}
    \\
	\subfloat[]{
        \includegraphics[width=0.98\linewidth]{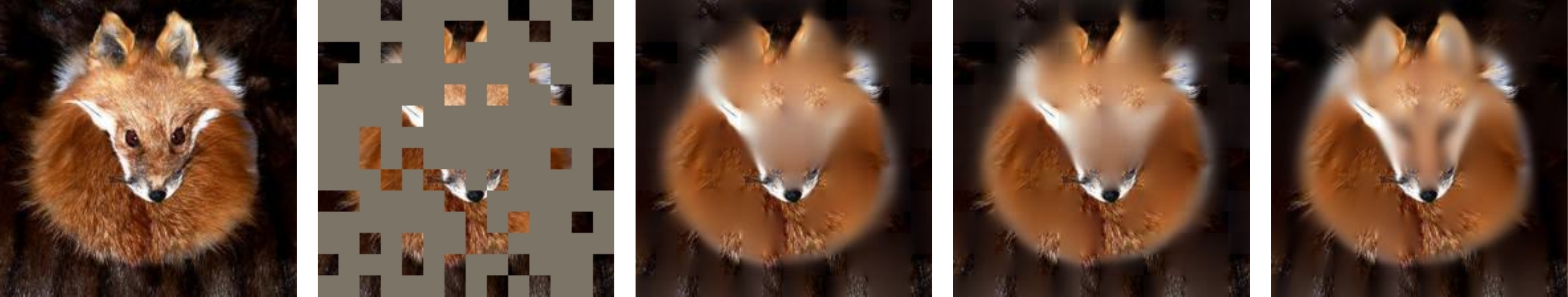}
    }
    \\
    \subfloat[]{
        \includegraphics[width=0.98\linewidth]{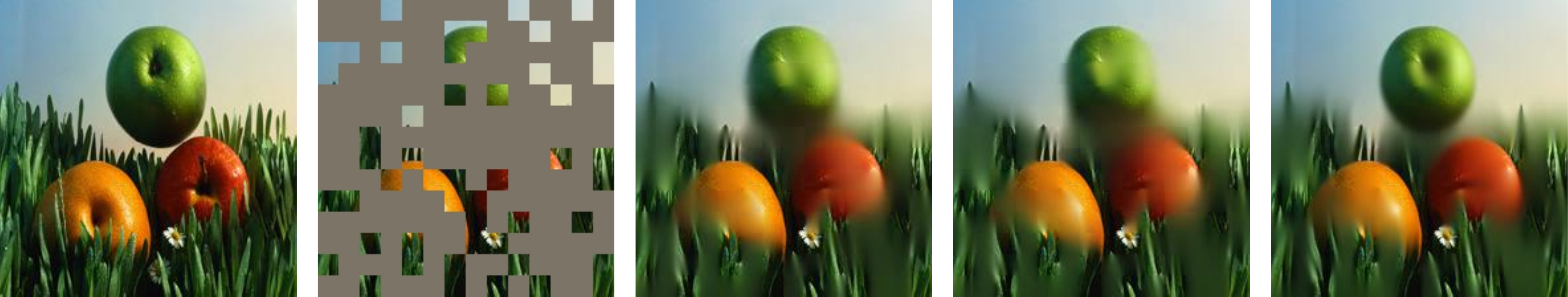}
    }
    \\
    \subfloat[]{
        \includegraphics[width=0.98\linewidth]{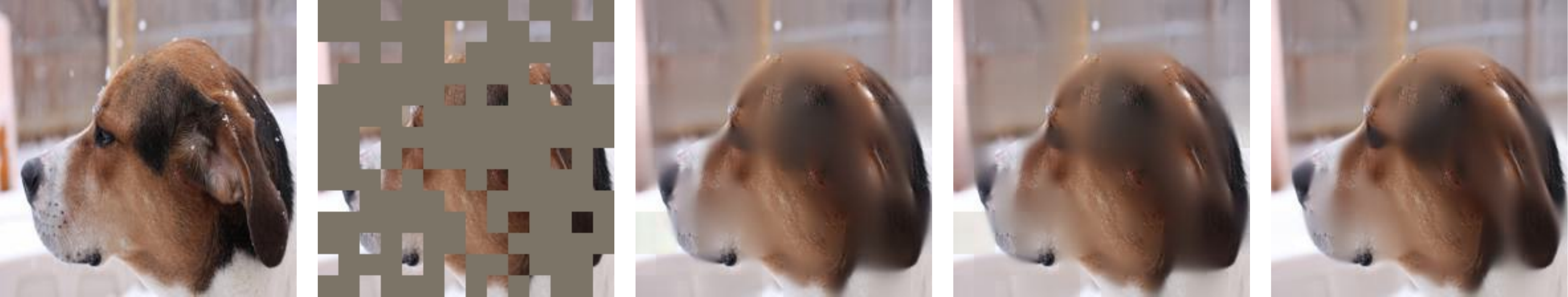}
    }
    \\
    \subfloat[]{
        \includegraphics[width=0.98\linewidth]{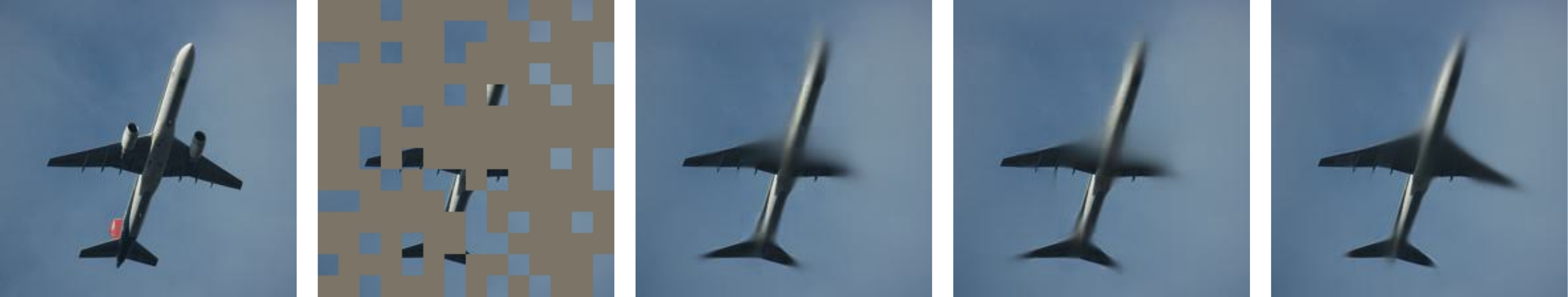}
    }
    \\
    \subfloat[]{
        \includegraphics[width=0.98\linewidth]{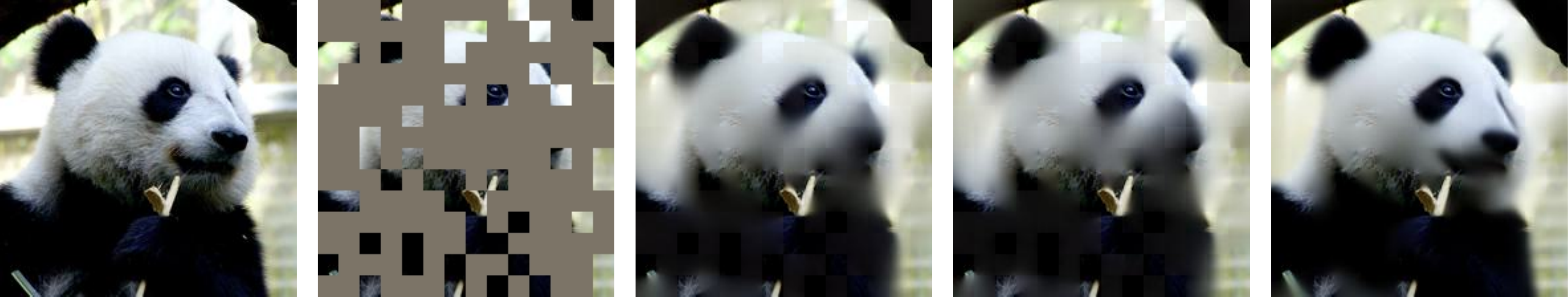}
    }
    \vspace{-10pt}
	\caption{MGP Reconstruction comparisons. Raw: input images. Masked: we fix all masks for intuitive comparisons. BAL-Base-400: ViT-Base-16 trained on ImageNet-BAL for 400 epochs. LT-Base-400: ViT-Base-16 trained on ImageNet-LT for 400 epochs. LT-Large-1600: ViT-Large-16 trained on ImageNet-LT for 1600 epochs. With the same training instance number and implementation settings, the ViT-B models trained with both LT and BAL datasets show comparable reconstruction ability. With the ImageNet-LT data, we can further get better reconstruction results with a bigger model and longer MGP epochs, as the column LT-Large-1600 shows.}
\end{figure*}

\end{document}